\documentclass[journal]{IEEEtran}
\usepackage[utf8]{inputenc}
\usepackage{csquotes}
\usepackage{hyperref}
\usepackage{graphicx}
\usepackage{multicol, multirow, bigstrut}
\usepackage{dirtytalk}
\ifCLASSOPTIONcompsoc
  \usepackage[caption=false,font=normalsize,labelfont=sf,textfont=sf]{subfig}
\else
  \usepackage[caption=false,font=footnotesize]{subfig}
\fi

\usepackage[bibstyle=ieee,
            citestyle=numeric-comp,
            minnames=1, maxnames=2,
            backend=biber,
            doi=false,isbn=false,
            url=true]{biblatex}
\AtEveryBibitem{\clearfield{month}}
\AtEveryCitekey{\clearfield{month}}

\usepackage{todonotes, comment, verbatim}
\usepackage{xcolor, soul}
\usepackage{enumitem}
\setlist{nolistsep}
\newcommand{\etal}{\textit{et al}.}
\newcommand{\eg}{e.g.,}
\newcommand{\ie}{i.e.,}
\newcommand{\github}{\href{https://github.com/gait-tech/gt.papers/tree/master/+ckf2019}{https://git.io/Je9VV}}

\usepackage{mdframed}
\usepackage{empheq}
\iffalse
    \global\mdfdefinestyle{mdchange}{%
    backgroundcolor=yellow, linewidth=0pt,%
    leftmargin=0pt,rightmargin=0pt,
    skipabove=0,skipbelow=0,
    innerleftmargin=0pt,innerrightmargin=0pt,
    innertopmargin=0pt,innerbottommargin=0pt
    }
    \DeclareRobustCommand{\changed}[1]{\hl{#1}}
    
\else
    \global\mdfdefinestyle{mdchange}{%
    linewidth=0pt,%
    leftmargin=0pt,rightmargin=0pt,
    skipabove=0,skipbelow=0,
    innerleftmargin=0pt,innerrightmargin=0pt,
    innertopmargin=0pt,innerbottommargin=0pt
    }
    \sethlcolor{white}
    \DeclareRobustCommand{\changed}[1]{\hl{#1}}
    
\fi

\usepackage{amsmath, amssymb, bm, units, arydshln}
\usepackage{nicefrac, xfrac}
\usepackage{mathtools}

\newcommand{\R}{\mathbb{R}}

\newcommand{\mat}[1]{\mathbf{#1}}
\renewcommand{\vec}[1]{\mathbf{\boldsymbol{#1}}}

\newcommand{\diag}{\mathbf{diag}}

\newcommand{\cs}[1]{#1}
\newcommand{\bs}[5]{\prescript{\cs{#1}}{#2}{#3}^{#4}_{#5}}
\newcommand{\bvraww}[5]{\prescript{\cs{#1}}{#2}{\vec{#3}}^{#4}_{#5}}

\newcommand{\bvUR}[5]{
    \ifthenelse{\equal{#4}{} \OR \equal{#2}{}}{
        \bvraww{#1}{}{#3}{\cs{#2}#4}{#5}
    }{
        ( \bvraww{#1}{}{#3}{\cs{#2}}{#5} )^{#4}
    }
}
\newcommand{\bv}[5]{
    \ifthenelse{\equal{#1}{W}}{
        \bvUR{}{#2}{#3}{#4}{#5}
    }{
        \bvUR{#1}{#2}{#3}{#4}{#5}
    }
}

\newcommand\bovermat[2]{%
  \makebox[0pt][l]{$\smash{\overbrace{\phantom{%
    \begin{matrix}#2\end{matrix}}}^{\text{#1}}}$}#2}

\newcommand{\qmult}{\otimes}
\newcommand{\quatUR}[5]{
    \ifthenelse{\equal{#4}{} \OR \equal{#2}{}}{
        \prescript{\cs{#1}}{}{\bm{#3}}^{\cs{#2}#4}_{#5}
    }{
        (\prescript{\cs{#1}}{}{\bm{#3}}^{\cs{#2}}_{#5})^{#4}
    }
}
\newcommand{\quatURR}[5]{
    \ifthenelse{\equal{#1}{W}}{
        \quatUR{}{#2}{#3}{#4}{#5}
    }{
        \quatUR{#1}{#2}{#3}{#4}{#5}
    }
}
\newcommand{\quatmeasUR}[4]{
    \quatURR{#1}{#2}{\breve{q}}{#3}{#4}
}
\newcommand{\quat}[4]{
    \quatURR{#1}{#2}{q}{#3}{#4}
}
\newcommand{\quatmeas}[4]{
    \quatURR{#1}{#2}{\breve{q}}{#3}{#4}
}
\newcommand{\quattilde}[4]{
    \quatURR{#1}{#2}{\tilde{q}}{#3}{#4}
}

\newcommand{\kfsp}[2]{\boldsymbol{\hat{#1}}^{-}_{#2}} % prediction update state
\newcommand{\kfsm}[2]{\boldsymbol{\hat{#1}}^{+}_{#2}} % measurement update state
\newcommand{\kfsc}[2]{\boldsymbol{\tilde{#1}}^{+}_{#2}} % constraint update state
\newcommand{\kfcp}[2]{\mat{#1}^{-}_{#2}} % prediction update Covariance matrix
\newcommand{\kfcm}[2]{\mat{#1}^{+}_{#2}} % measurement update Covariance matrix
\newcommand{\kfcc}[2]{\mat{\tilde{#1}}^{+}_{#2}} % measurement update Covariance matrix
\newcommand{\dt}{\Delta t}

\soulregister\cite7
\soulregister\ref7
\soulregister\pageref7
\soulregister\eqref7

\title{Estimating Lower Limb Kinematics \\ using a Reduced Wearable Sensor Count}
\author{Luke Sy \emph{Student Member, IEEE}, Michael Raitor \emph{Student Member, IEEE}, \\
    Michael Del Rosario \emph{Member, IEEE}, Heba Khamis \emph{Member, IEEE}, Lauren Kark, \\
    Nigel H. Lovell \emph{Fellow, IEEE}, Stephen J. Redmond \emph{Senior Member, IEEE}}
\date{20 June 2018}
\IEEEaftertitletext{\vspace{-2\baselineskip}}

\begin{document}
	\maketitle
	\bstctlcite{IEEEexample:BSTcontrol}
		
	\begin{abstract}
	    % how does the algorithm work abstractly speaking
	    \emph{Goal:} This paper presents an algorithm for accurately estimating pelvis, thigh, and shank kinematics during walking using only three wearable inertial sensors. 
	    \emph{Methods:} The algorithm makes novel use of a constrained Kalman filter (CKF). 
	    % what are the components of the algorithm
	    The algorithm iterates through the prediction (kinematic equation), measurement (pelvis position pseudo-measurements, zero velocity update, flat-floor assumption, and covariance limiter), and constraint update (formulation of hinged knee joints and ball-and-socket hip joints).
	    % describe the overall results
	    \emph{Results: } Evaluation of the algorithm using \changed{an optical motion capture}-based sensor-to-segment calibration on nine participants ($7$ men and $2$ women, weight $63.0 \pm 6.8$ kg, height $1.70 \pm 0.06$ m, age $24.6 \pm 3.9$ years old), with no known gait or lower body biomechanical abnormalities, who walked within a $4 \times 4$ m$^2$ capture area shows that it can track motion relative to the mid-pelvis origin with mean position and orientation (no bias) root-mean-square error (RMSE) of $5.21 \pm 1.3$ cm and $16.1 \pm 3.2^\circ$, respectively. The sagittal knee and hip joint angle RMSEs (no bias) were $10.0 \pm 2.9^\circ$ and $9.9 \pm 3.2^\circ$, respectively, while the corresponding correlation coefficient (CC) values were $0.87 \pm 0.08$ and $0.74 \pm 0.12$.
	    \emph{Conclusion: } The CKF-based algorithm was able to track the 3D pose of the pelvis, thigh, and shanks using only three inertial sensors worn on the pelvis and shanks.
	    \emph{Significance: } Due to the Kalman-filter-based algorithm\mbox{’}s low computation cost and the relative convenience of using only three wearable sensors, gait parameters can be computed in real-time and remotely for long-term gait monitoring. Furthermore, the system can be used to inform real-time gait assistive devices.
	\end{abstract}


	\section{Introduction}
	% Gait analysis and its motivation
	Gait analysis (GA), part of the study of human movement, is a valuable clinical tool that can help diagnose movement disorders and assess surgical outcomes \cite{Shull2014}. % \cite{Simon2004, Iosa2016, Shull2014}
	    Its value has been demonstrated in studies related to: 
    	    knee and hip osteoarthritis \cite{Ornetti2010}; 
    	    falls risk \cite{Hausdorff2001};
    	    % \cite{Maki1997, Hausdorff2001}; 
    	    and children with cerebral palsy \cite{VanDenNoort2013}.
    	    % and Parkinson's disease \cite{Sofuwa2005}.
	    % and outcome assessment of total knee replacement and total hip arthroplasty \cite{Jolles2012, Reininga2013}. 
	    Such studies quantitatively analyze gait parameters which are typically categorized as either spatiotemporal (\eg{} speed and stride length), kinematic (\eg{} knee flexion), or kinetic (\eg{} joint moments) parameters \cite{Shull2014}. % \cite{Shull2014, Iosa2016}
	GA is also used in interactive and possibly unsupervised rehabilitation, where measured gait parameters can drive real-time feedback (\eg{} \cite{Shull2010}).

    Gait parameters can be objectively measured by human motion capture systems (HMCSs). HMCSs can be organized into two broad groups: non-wearable and wearable.
	% Nonwearable HMCS
	Non-wearable systems, predominantly camera-based, can estimate position with up to millimeter accuracy, if well-configured and calibrated.
	    \changed{Optical motion capture (OMC) systems} are currently the industry standard for motion capture in the laboratory setting.
        However, the capture volume is usually restricted to tens of cubic meters, thereby limiting the duration and range over which a person's gait can be assessed. 
        The unfamiliarity or unusualness of this environment may also lead the person to move in manners different to their normal gait \cite{Dingwell2001}.
            % For example, Dingwell \etal{} showed that treadmill walking significantly reduces variability in lower body motion compared to normal overground walking \cite{Dingwell2001}.
    % Wearable HMCS
    Wearable systems primarily use inertial measurement units (IMUs) which measure the linear acceleration, angular velocity, and magnetic field.
        % (\eg{} MOVEN Xsens from Xsens technologies B.V., Enschede, the Netherlands)
        Recent technological advancements have dramatically miniaturized IMUs, making it arguably the most promising technology for long-term and continuous tracking of human movement during everyday life \cite{Tedesco2017}.
        % Tracking gait in the subject's natural environment using wearable sensors improves the likelihood of obtaining representative measurement of their gait, while also facilitating long-term monitoring such that the progression of a disorder or an intervention can be tracked over timescales spanning days to years.
        Furthermore, wearable systems are typically cheaper and easier to configure, enabling the capture of large number of participants in research studies (\eg{} \cite{Tudor-Locke2012}). 
        % For example, Tudor-Locke \etal{} was able to study the relation of sex, age, and BMI to peak stepping cadence on more than 3,000 participants using accelerometers worn on the belt for seven consecutive days \cite{Tudor-Locke2012}.
    
    Wearable systems can measure gait parameters derived from one (\eg{} one foot) or multiple body segments.
        This paper will focus on HMCSs that monitor the kinematics of the lower body (\eg{} hip and knee angles).
        Commercial wearable HMCSs attach one sensor per body segment (OSPS) (\ie{} five sensors to track the pelvis, thighs, and shanks) \cite{Roetenberg2009, LEGSys-BioSensics}.
            % \cite{Roetenberg2009,Tadano2013,Zihajehzadeh2017,LEGSys-BioSensics}.
            A sensor-to-segment calibration relates the IMU frame with the body segment frame, and then each
            body segment's 3D orientation, angular rate, and acceleration are estimated based on the assumption that the sensor is rigidly attached to the body segment it is monitoring.
                The sensor-to-segment calibration can be achieved through either careful manual placement, use of a relatively tedious set of calibration movements, or an automatic online calibration method which leverages assumptions about joint kinematics \cite{Roetenberg2009, Laidig2017}.
                %Based on this assumption, one can obtain that body segment's 3D orientation, rate of rotation, and acceleration.
            In general, position is inferred through double integration of the measured acceleration and/or by vector addition of segment axes where segments make a kinematic chain,
                while orientation is inferred through integration of measured angular rate.
            In practice, position estimates will experience severe drift if not provided with at least an intermittent external reference of position or velocity, 
                and orientation estimates will drift if not corrected using an accelerometer (using acceleration due to gravity to correct the pitch and roll) or a magnetometer (using magnetic north to correct yaw) \cite{DelRosario2016a, DelRosario2018}.
            Acceptable measurements vary between clinical applications. 
            However, McGinley et.al. believe that in most common clinical situations an RMSE of $2^\circ$ or less is considered excellent, and RMSE between $2^\circ$ and $5^\circ$ is acceptable but may require consideration in data interpretation \cite{McGinley2009}. 
            OSPS typically have acceptable results ($<5^\circ$ RMSE no bias and adjusted for repeatability) \cite{Al-Amri2018}.
            The number of IMUs required to be attached by OSPS configurations might be considered too cumbersome or inconvenient for routine daily use, and would currently be considered too expensive by the consumer market, costing tens of thousands of US dollars for a state-of-the-art full-body system.
        We expect that a reduced-sensor-count (RSC) configuration, where sensors are placed on a subset of the total number of body segments, will improve user comfort while also reducing setup time and system cost.
            However, reducing the number of sensors inherently reduces available kinematic information, from which body kinematics are inferred.
    %Learning from the commercial success of fitness trackers, other than being accurate and portable, wearable systems must integrate seamlessly into the user's everyday life (\ie{} be intuitive, comfortable, unobtrusive, stigma free, and aesthetically pleasing) \cite{Chan2012, Tedesco2017, Chaudhuri2017}.
        % It needs to be unnoticeable due to the belief among elderly that such devices symbolize loss of independence \cite{Chaudhuri2017}.
        %; however, given the mechanical linkage between body segments, the rate at which information is surrendered with each fewer sensor used is not expected to be linear.
            % Hence, we are motivated to explore RSC configurations which retain sufficient information to allow the reconstruction of body kinematics.
            In the following, we briefly review works which infer body kinematics for RSC configurations, arranged under data-driven and model-based approaches.
            
    \subsection{Approaches to kinematic inference for RSC systems}
    \subsubsection{Data-driven kinematic inference}
    % How current sparse WHMCS works (data driven)
    These approaches attempt to reconstruct the missing data by comparing the data obtained to an existing motion database (DB) or by some learned model \cite{Tautges2011, Wouda2016, Huang2018a}. 
        It assumes that kinematic parameters are correlated, such that a pattern of available measurements allows the missing kinematic parameters to be inferred from previously acquired distributions representing a broad range of common human movements. 
        Tautges \etal{} searched for the best motion from their DB through a nearest-neighbor search of poses at multiple time steps \cite{Tautges2011}. 
            % The algorithm accepted acceleration data from IMUs worn on the ankles and wrist as inputs.
            % around 6 cm? the paper didn't write the result directly....
        Wouda \etal{} utilized five IMUs attached to the pelvis, arms, and legs, and inferred the orientation of the body segments without sensors attached using a k-nearest-neighbour algorithm and artificial neural network  \cite{Wouda2016}. 
            % The IMUs on the arms and legs were tested at different locations (\eg{} upper arm, hands, thigh and shank)
            % As inference was done only at each time step, it was computationally efficient, but at the risk of having a non-smooth reconstructed motion.
            % The algorithm was applied at each time step, so temporal information was not utilised.
            % This method is more computationally efficient as only the information per time step is considered, but at the cost of having a non-smooth reconstructed motion.
            % average joint position error of approximately 7 cm, and average joint angle error of 7◦.
        % Mousas used a hierarchical multivariate hidden Markov model to reconstruct body pose from just one IMU \cite{Mousas2017}. 
            % Mousas tested the algorithm under five different IMU sensor placements (\eg{} attached to the right hand, right forearm, right foot, right knee, and pelvis).
            % 4.53 cm but only tested on 6000 frames (that's around 60 or 200 sec of recording, the unsureness came from ambiguity)
        \changed{Huang \mbox{\etal{}} developed a deep neural network with a bi-directional recurrent neural network architecture that reconstructs full body pose using six IMUs located on the head, wrists, pelvis, and ankles \cite{Huang2018a}.}
    Data-driven approaches have been shown to efficiently reconstruct realistic motions making them useful for animation-related applications \cite{Tautges2011, Wouda2016, Huang2018a}. %\cite{Tautges2011, Wouda2016, Mousas2017}
        However, these approaches naturally have a bias toward motions already contained in the DB, inherently limiting their use in pathological gait monitoring.
    % However, its use in clinical applications can be difficult as GA needs to be done on the subject's movement. Data-driven approaches may find it difficult to classify if the capture is the subject's actual movement to obtain accurate insight regarding the subject's health.
    %    Hence, an additional step of segregation of which part of the reconstructed motion came from the subject's motion or from the DB is needed.
    % However, using data-driven approaches in clinical applications can be difficult due to the tendency of these approaches to output the inferred body poses that are similar to examples which appear in the database, but which may not be physically plausible for the subject under observation. 
    %    Furthermore, as motion databases tend to contain only common motions (\ie{} motion from healthy subjects), the motion of interest (\ie{} motion from subjects with abnormal gait) might not be represented in the database, leading ultimately to erroneous clinical evaluation.
    
    \subsubsection{Model-based kinematic inference}
    % How current reduced WHMCS works (model based)
    These approaches attempt to reconstruct body motion using kinematic and biomechanical models. % , typically modelling the human body as linked, rigid body segments.
        % Salarian \etal{} modeled each leg as a double pendulum, and inferred the thigh\rq{}s sagittal orientation using a linear regression model from measurements of the shank\rq{}s sagittal orientation \cite{Salarian2013}.
            % , the latter obtained by integrating the angular rate measured using gyroscopes (corrected by an accelerometer) \cite{Salarian2013}.
            % Fourier harmonics were extracted from the shank orientation as input features to train a linear regression model which estimated the Fourier harmonics of the thigh orientation.
            % Finally, spatiotemporal parameters (stride length and velocity) and range of rotation were calculated from the thigh and shank sagittal orientations.
        Hu \etal{} modeled each leg as three linked segments confined in the sagittal plane.
            It tracked pelvis and ankle positions from four IMUs (two at pelvis and one for each foot) while making velocity assumptions at certain  gait cycle stages, 
            and calculated hip, knee, and ankle angles using inverse kinematics \cite{Hu2015}. 
            However, the system could only track motion in the sagittal plane, which can cause issues when tracking activities of daily living (ADLs) such as side or diagonal steps.
        Marcard \etal{} modeled the body as a kinematic chain of rigid segments linked by 24 ball and socket joints \cite{Marcard2017}.
            A window-based optimization algorithm then calculates the model pose that best matches the orientation and acceleration measurements from six IMUs located on the head, wrists, pelvis, and ankles, while satisfying multiple joint angle constraints.
            % Although the system performed well ($3.9 \pm 4.0$ cm position error),
            % $13.32 \pm 10.13^{\circ}$
            % it took 7.5 minutes to process 17 seconds of movement on an i7 3.5 GHz single core computer.
            However, it has long computation time, and hence only suitable for offline analysis applications.
            % the system cannot be used in conjunction with gait assistive devices that can correct the user\rq{}s gait in real time (\eg{} \cite{Shull2010, Llorens2015}).
    
    % novelty
    \subsection{Novelty}
    This paper describes a novel model-based algorithm based on a constrained Kalman filter (CKF) to estimate lower body kinematics using a RSC configuration of IMUs.
        % A Kalman Filter (KF) outputs model state estimates at every time step, and is therefore a suitable algorithm for real-time applications.
        A Kalman Filter (KF) is the optimal state estimator for linear dynamic systems. 
            It is capable of handling uncertainty in measurements (\eg{} IMU measurements) and indirect observations (\eg{} step detection and biomechanical constraints) \cite{simon2006optimal}.
            It also outputs model state estimates at every time step, and is therefore a suitable algorithm for real-time applications. 
    This design was motivated by the need to develop a gait assessment tool using as few sensors as possible, ergonomically-placed for comfort, to facilitate long-term monitoring of lower body movement.
    % IMU and optical motion capture data collected for walking motions under RSC configuration.
	
	\section{Algorithm Description}
	    % problem definition
	    
\begin{figure}[htbp]
	        \centering
	        \includegraphics[width=\linewidth]{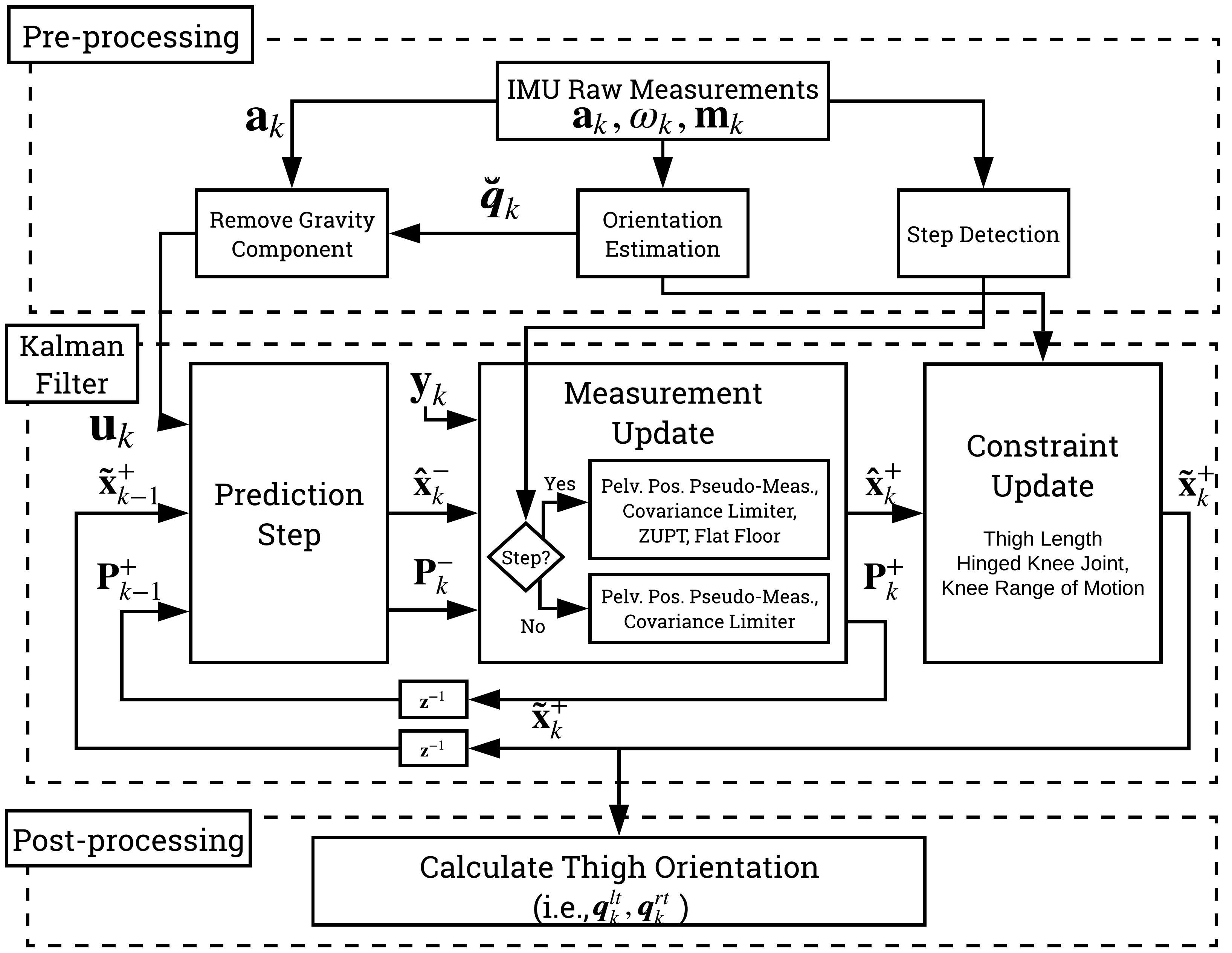}
	        \caption{Algorithm overview which consists of pre-processing, KF based state estimation, and post-processing. Pre-processing calculates the inertial acceleration of each IMU (fixed on various body segments), body segment orientation, and step detection from raw acceleration, $\vec{a}_{k}$, angular velocity, $\vec{\omega}_{k}$, and magnetic north heading, $\vec{h}_{k}$, measured by the IMU. The KF-based state estimation consists of a prediction (kinematic equation), measurement (pelvis position pseudo-measurements, covariance limiter, intermittent zero-velocity update, and flat-floor assumption), and constraint update (thigh length, hinge knee joint, and knee range of motion). Post-processing calculates the thigh orientation, $\bm{q}_{lt}$ and $\bm{q}_{rt}$.}
	        \label{fig:algo-overview}
	    \end{figure}

    	The overall objective is to estimate the orientation of the pelvis, thighs, and shanks with respect the world frame, $W$ (see Fig. \ref{fig:body-skeleton-sparse-ckf}); 
    	    their orientation completely defines their relative positions, since they form a mechanical linkage through hip and knee joints. 
	    At each time step, we attempt to predict the location of the shanks and pelvis in 3D space through double integration of their linear 3D acceleration, as measured by the IMUs attached to these three body segments; 
	        note, a pre-processing step fuses the IMU accelerometer, gyroscope, and magnetometer readings to estimate the IMU (and hence associated body segment) orientation so that gravitational acceleration can be removed, giving 3D linear acceleration of each IMU (and associated body segment) in the world frame. 
	        To mitigate positional drift due to sensor bias and errors in the double integration of acceleration, the 3D velocity and the height above the floor of the ankle is zeroed whenever a footstep is detected on that leg. 
	        This zeroing of ankle velocity and correction of ankle position is implemented within the KF framework by making a pseudo-measurement; 
	        to control the otherwise ever-growing error covariance for the pelvis and ankle positions, a pseudo-measurement equal to the current position state estimate with a fixed covariance is made. 
	        Next, we make two novel contributions which take us beyond state-of-the-art: 
	            (1) We mitigate position drift in the pelvis, caused by error accumulation during double integration of acceleration, by making a noisy pseudo-measurement of 3D position of the pelvis as being the length of the unbent leg(s) above the floor, with an XY position of the pelvis being mid-way between the two ankle locations; in practical terms, to say we make a noisy pseudo-measurement means that we assume there is large noise covariance associated with this pseudo-measurement, which we set empirically to represent our uncertainty in this assumption. This effectively puts a soft constraint on the 3D pelvis position relative to the two ankles. 
	            (2) We ensure that the positions of the shanks and pelvis are such that they obey the assumed biomechanical constraints of the lower body, being that the hip joints are ball-and-sockets and the knee joint is a hinge with one degree of freedom and a limited range of motion (ROM). 
	            These constraints result in three equations (one of which is non-linear) in the XYZ position coordinates of the shanks and pelvis which ensure that: 
	                (i) the knee-to-hip distance is equal to the thigh length; 
	                (ii) long axis of the thigh is orthogonal to the mediolateral axis of the shank (\ie{} knee is a hinge), and; the knee joint angle is within a biomechanically-realistic range. 
                Again, these constraints can be implemented in the KF framework as pseudo-measurements, although the non-linear constraint equation must be linearized first; 
                furthermore, since we are linearizing non-linear constraint equations, it is necessary to iterate this pseudo-measurement procedure at each time step until these non-linear constraints are satisfied to within some tolerance. 
                A final important point is that throughout the entire algorithm the orientation estimated using the IMU during the pre-processing stage is taken to be free of error, which is obviously untrue and will be the topic of future work. 
                The three stages of the algorithm (i.e., prediction, measurement, and constraint updates) are shown in Fig. \ref{fig:algo-overview}.
	    
	    In the next subsection, the chosen model and notation are defined, followed by an illustration of the selected IMU placement, and a detailed explanation of the algorithm.
	    
	\subsection{Physical model and mathematical notation}
	    % joint position
        The model makes use of several important joints and points of the lower body; namely, the mid-pelvis (taken as mid-point between hip joints), hip joints, knee joints, and ankles (\ie{} $\mathbb{P}$ = $\{ \bv{W}{mp}{p}{}{}, \bv{W}{lh}{p}{}{}, \bv{W}{rh}{p}{}{}, \bv{W}{lk}{p}{}{}, \bv{W}{rk}{p}{}{}, \bv{W}{la}{p}{}{}, \bv{W}{ra}{p}{}{} \}$).
        $\bv{D}{C}{p}{}{} \in \R^3$ denotes the 3D position of joint $\cs{C}$ with respect to frame $\cs{D}$. 
        If frame $\cs{D}$ is not specified, assume reference to the world frame, $\cs{W}$.
        
        % segment orientation
        Additionally, the lower body is comprised of several segments, namely, the pelvis, thighs, and shanks.
            Each segment has its own length, measured joint-to-joint (\ie{} $\mathbb{L}$ = \{$d^{\cs{p}}, d^{\cs{lt}}, d^{\cs{rt}}, d^{\cs{ls}}, d^{\cs{rs}} \}$), 
            and orientation, represented using quaternions (\ie{} $\mathbb{O}$ = $\{ \quat{W}{p}{}{}, \quat{W}{lt}{}{}, \quat{W}{rt}{}{}, \quat{W}{ls}{}{}, \quat{W}{rs}{}{} \}$).
        Quaternion $\quat{D}{C}{}{} \in \R^4$ denotes the orientation of frame $\cs{C}$ with respect frame $\cs{D}$ and will be represented as bold-faced and italicized.
        If frame $\cs{D}$ is not specified, assume reference to the world frame $\cs{W}$.
        Note that unit quaternions can be interchangeably expressed as a rotation matrix, \ie{} 
            $\quat{W}{S}{}{} \Longleftrightarrow
                \begin{bmatrix}
                    \bv{W}{S}{r}{}{x} & \bv{W}{S}{r}{}{y} & \bv{W}{S}{r}{}{z}
                \end{bmatrix}$
        where $\bv{W}{S}{r}{}{x}$, $\bv{W}{S}{r}{}{y}$, and $\bv{W}{S}{r}{}{z} \in \R^3$ are the basis vectors of the segment's local frame, $\cs{S}$, with respect to the world frame, $\cs{W}$.
        Details on how to convert between quaternion and rotation matrix representations of 3D orientation can be found in \cite{Kuipers1999}.
	    
	\subsection{IMU Setup}
\begin{figure}[htbp]
	        \centering
	        \includegraphics[width=0.35\textwidth]{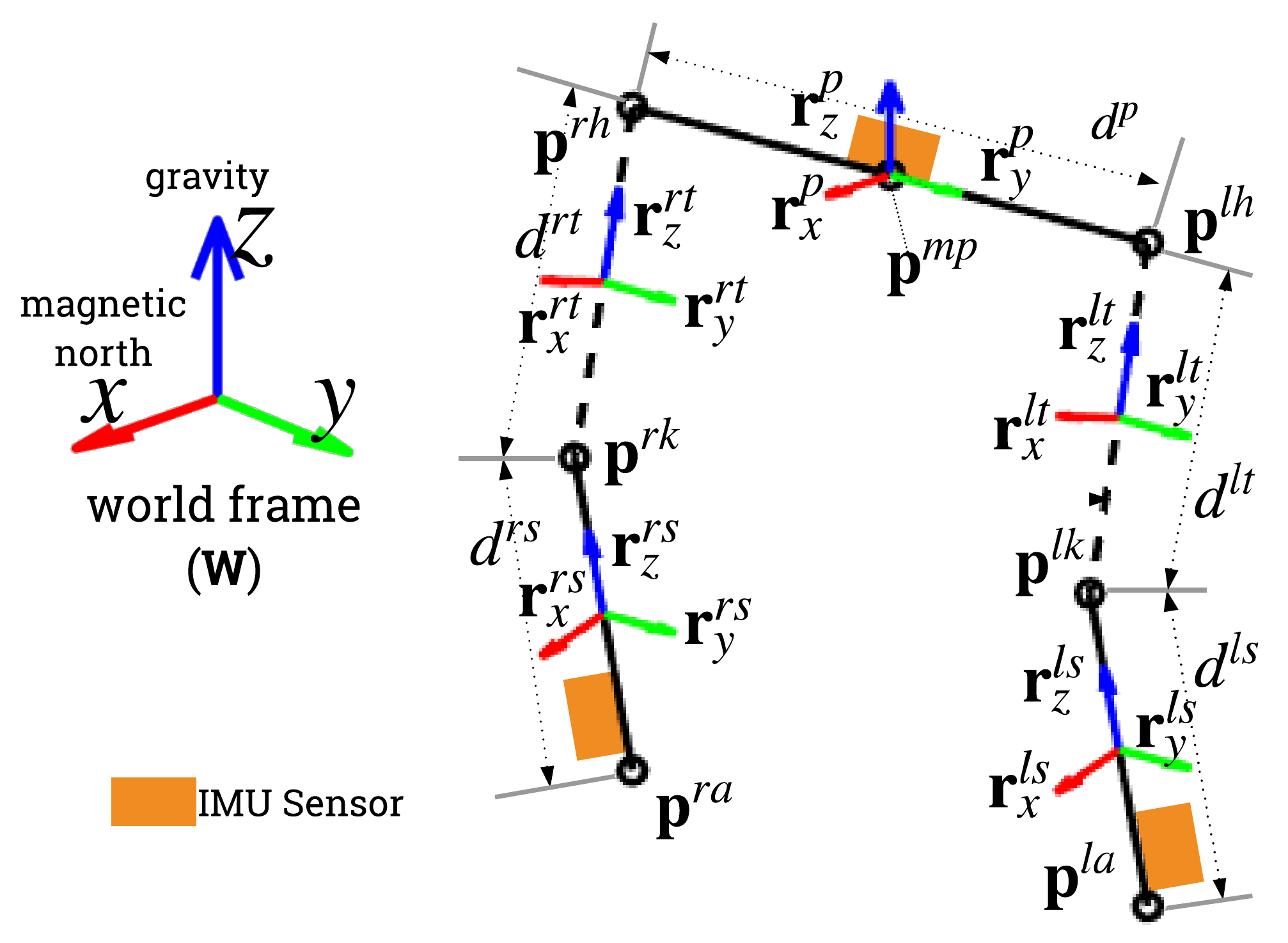}
	        \caption{Physical model of the lower body used by the algorithm. The circles denote the joint positions $\mathbb{P}$. The solid lines denote instrumented body segments, whilst the dashed lines denote segments without IMUs attached. Estimating the orientation and position of the uninstrumented thighs is the main contribution of this paper.}
	        \label{fig:body-skeleton-sparse-ckf}
	    \end{figure}

	    The system will be instrumented with an IMU at each of the three locations: (1) the sacrum, approximately coincident with the mid-pelvis; (2 \& 3) the left and the right shanks, just above the ankles, as shown in Fig. \ref{fig:body-skeleton-sparse-ckf}.
	    Note that the noise from IMU measurements (\eg{} acceleration) when stationary can be approximated by a normal distribution  \mbox{\cite[Ch. 2]{Kok2017}}, but more sources of error, not necessarily normally distributed as assumed by the KF, comes into play during dynamic motion \cite{woodman2007introduction}.
	        Nevertheless, several publications have successfully tracked pose while approximating IMU measurements as a normal distribution \cite{Roetenberg2009, DelRosario2018}.
	    
    \subsection{Pre-processing}
        Pre-processing involves the calculation of orientation and inertial acceleration of instrumented body segments from raw IMU measurements, and step detection for both feet. 
        The information obtained is then used by the KF state estimator.
    
    \subsubsection{Body segment orientation} \label{sec:body-seg-ori}
        
            The reference frame of each IMU may not be coincident with that of the attached body segment. 
            This misalignment is corrected using a sensor-to-segment realignment similar to \mbox{\cite[Sec. II-B Eq. (1)]{Roetenberg2009}} where the orientation of the instrumented body segment at time step $k$,
    		\ie{} $\mathbb{IO}$ = $\{ \quatmeas{W}{p}{}{k}, \quatmeas{W}{ls}{}{k}, \quatmeas{W}{rs}{}{k} \}$, 
    		is calculated from (i) measured sensor orientation of sensor frame, $S$, in the world frame, $W$, and (ii) a known body segment orientation (\eg{} measured by \changed{an OMC} system).
    \subsubsection{Inertial acceleration of IMUs in the world frame} \label{sec:free-body-acc}
         is due to body movement, as distinct from gravitational acceleration which is measured by the IMU's accelerometer. 
        	It is calculated by subtracting acceleration due to gravity and then expressing the acceleration of the instrumented body segment in the world frame similar to \mbox{\cite[Eq. (3)]{Roetenberg2009}}.
        \begin{comment}
        Inertial acceleration is due to body movement, as distinct from gravitational acceleration which is also present in the IMU's accelerometer measurements.     
        It is calculated by expressing the acceleration of the instrumented body segment in the world frame, after which acceleration due to gravity, $\bv{W}{}{g}{}{}$, must be subtracted, as shown in Eq. \eqref{eq:inertial-accel-global-frame}.
        Note that $\bv{S}{}{a}{}{k}$ is the raw accelerometer measurement of the instrumented body segments,
            and $\mathring{\makebox[\widthof{q}]{}}$ denotes augmenting zero to the column vector (\ie{} $\mathring{\vec{\tau}}=\begin{bmatrix} 0 & \vec{\tau}^T \end{bmatrix}^T$).
        \begin{align}
            \bvUR{W}{S}{\mathring{a}}{}{k} &= (\quatmeasUR{W}{S}{}{k} \qmult \bvUR{S}{}{\mathring{a}}{}{k} \qmult \quatmeasUR{W}{S}{-1}{k}) - \bv{W}{}{\mathring{g}}{}{} \label{eq:inertial-accel-global-frame}
        \end{align}
        \end{comment}
        
    \subsubsection{Step Detection (SD)} \label{sec:step-detection}
        detects if the left or right foot is in contact with the ground. 
        It can be used to gain knowledge about the velocity and position of the ankle, treated as an intermittent measurement in the KF. 
        This component can use any third-party SD algorithm. 
        A review on different SD algorithms can be found in \cite{Skog2010}.
        
	\subsection{System, measurement, and constraint models}
	    The state estimator is based upon a CKF, which combines a system model (Section \ref{sec:pred-update}), a measurement model (Section \ref{sec:meas-update}), and a kinematic constraint model (Section \ref{sec:const-nonlin-update}). The interaction between these models is illustrated in Fig. \ref{fig:algo-overview}. Here, we first introduce these models before presenting the CKF.
    	    \begin{gather}
    	        \vec{x}_{k} = \mat{F} \vec{x}_{k-1} + \mat{G} \vec{u}_{k-1} + \vec{w}_{k-1} \label{eq:pred-update} \\
    	        \vec{y}_{k} = \mat{H}_{k} \vec{x}_{k} + \vec{v}_{k}, \qquad 
    	        \mat{D}_{k} \vec{x}_{k} = \vec{d}_{k} \label{eq:meas-const-update}
            \end{gather}
        where $k$ is the time step; $\vec{x}_{k}$ is the state vector; $\vec{y}_{k}$ is the measurement vector; $\vec{w}_{k}$ and $\vec{v}_{k}$ are zero-mean process and measurement noise vectors with covariance matrices $\mat{Q}_{k}$ and $\mat{R}_{k}$, respectively (denoted as $\vec{w}_{k} \sim (\vec{0}, \mat{Q}_{k})$ and $\vec{v}_{k} \sim (\vec{0}, \mat{R}_{k})$; $\mat{F}$, $\mat{G}$, and $\mat{H}_{k}$ are the state transition, input, and measurement matrices, respectively; and $\mat{D}_{k} \vec{x}_{k} = \vec{d}_{k}$ are the equality constraints that state $\vec{x}_{k}$ must satisfy.
        \changed{Eqs. \eqref{eq:pred-update} and \eqref{eq:meas-const-update} are the standard system model used in CKF. A more detailed description can be found in }\cite{simon2006optimal}.
        The state variables in $\vec{x}_{k}$ models the position and velocity of the instrumented body segments (\ie{} the $18 \times 1$ vector $\vec{x}_{k}$ = $\big[\bv{W}{mp}{p}{T}{k}$, $\bv{W}{la}{p}{T}{k}$, $\bv{W}{ra}{p}{T}{k}$, $\bv{W}{mp}{v}{T}{k}$, $\bv{W}{la}{v}{T}{k}$, $\bv{W}{ra}{v}{T}{k} \big]^{T}$).
        % \changed{Note that a state representation including acceleration (\ie{} constant acceleration model) is also valid, but the current $18 \times 1$ vector representation was chosen because the authors find it more intuitive and it requires less dimensions \cite{Kok2017}.}
        
    \subsection{Constrained Kalman filter (CKF)}
        The \textit{a priori} (predicted), \textit{a posteriori} (updated using measurements), and constrained state (satisfying the constraint equation) for time step $k$ are denoted by $\kfsp{x}{k}$, $\kfsm{x}{k}$, and $\kfsc{x}{k}$, respectively. The KF state error \textit{a priori} and \textit{a posteriori} covariance matrices are denoted as $\kfcp{P}{k}$ and $\kfcm{P}{k}$, respectively.
       % Note that an implementation of this CKF algorithm will be made available at: \github{}.
       An implementation will be made available at: \github{}.
	    % the calculation of orientation from the tracked state (Section \ref{sec:state-to-ori}). 

    \subsubsection{Prediction step} \label{sec:pred-update}
        % The prediction step 
        estimates the kinematic states at the next time step. This prediction does not respect the kinematic constraints of the body, so joints may become dislocated after this prediction step. 
        The input is the inertial acceleration, expressed in the world frame, as described in Section \ref{sec:free-body-acc} (\ie{}  $\vec{u}_{k} = \begin{bmatrix} \bv{W}{mp}{a}{T}{k} & \bv{W}{la}{a}{T}{k} & \bv{W}{ra}{a}{T}{k} \end{bmatrix}^T$). 
        The matrices $\mat{F}$, $\mat{G}$, and $\mat{Q}$ are defined following the kinematic equations of motion for the pelvis and ankles as shown below; that is velocity is the integral of acceleration, and position the integral of velocity:
        % \begin{empheq}[box=\myyellowbox]{gather}
        \begin{gather}
            \mat{F} = \begin{bmatrix}
                    \mat{I}_{9\times9} & \dt \mat{I}_{9\times9} \\
                    \mat{0}_{9\times9} & \mat{I}_{9\times9}
                \end{bmatrix}, \quad
            \mat{G} = \begin{bmatrix}
                    \frac{\dt^2}{2} \mat{I}_{9\times9} \\
                    \dt \mat{I}_{9\times9}
                \end{bmatrix} \\
            \mat{Q} = \mat{G} \text{ } \diag(\bv{}{}{\sigma}{2}{acc}) \text{ } \mat{G} ^T 
        \end{gather}
        % \end{empheq}

        where $\bv{}{}{\sigma}{2}{acc}$ is a $9 \times 1$ vector of variances in the accelerometer measurements (assumed equal variance for all accelerometry signals), 
        % $\bv{}{}{\sigma}{2}{wori}$ is a $12 \times 1$ vector of variances in the measured quaternion components (again, all assumed equal variance), 
        and $\dt$ is the sampling interval.
        
        The \textit{a priori} state estimate, $\kfsp{x}{k}$, and state estimate error covariance, $\kfcp{P}{k}$, are calculated following the standard KF equations \cite{Kalman1960}:
        \begin{align}
            \kfsp{x}{k} = \mat{F} \kfsc{x}{k-1} + \mat{G} \vec{u}_{k}, & \qquad
            \kfcp{P}{k} = \mat{F} \kfcm{P}{k-1} \mat{F}^T + \mat{Q}
        \end{align}
        
    \subsubsection{Measurement update} \label{sec:meas-update}
        % The measurement update 
        estimates the next state by: (i) enforcing soft pelvis position constraints based on ankle $x$ and $y$ position, and initial pelvis height, and by; (ii) utilizing zero ankle velocity and flat floor assumptions whenever a footstep is detected.
            For the covariance update, there is an additional step to limit the \textit{a posteriori} covariance matrix elements from growing indefinitely and from becoming badly conditioned.

        Firstly, the soft mid-pelvis position constraints encourage: 
            (i) the pelvis $x$ and $y$ position to approach the average of the left and right ankle $x$ and $y$ positions; 
            and (ii) the pelvis $z$ position to be close to the initial pelvis $z$ position as time $k=0$ (\ie{} standing height $z_{p}$).
            These constraints are implemented by $\mat{H}_{mp}$ and $\vec{y}_{mp}$ as shown in Eqs. \eqref{eq:H-pelv-k} and \eqref{eq:y-pelv-k} with measurement noise variance $\bv{}{}{\sigma}{2}{mp}$ ($3 \times 1$ vector).
            % The soft constraints encourage: (i) the pelvis $x$ and $y$ position to approach the average of the left and right ankle $x$ and $y$ positions; and (ii) the pelvis $z$ position to be close to the initial pelvis $z$ position as time $k=0$ (\ie{} standing height $z_{p}$).
        Secondly, if a left step is detected, the left ankle velocity is encouraged to approach zero, and the left ankle $z$ position to be close to the floor level, $z_{f}$.
            These assumptions are implemented using $\mat{H}_{ls}$ and $\vec{y}_{ls}$ as shown in Eq. \eqref{eq:H-y-lstep-k} with measurement noise variance $\bv{}{}{\sigma}{2}{ls}$ ($4 \times 1$ vector).
            % In plain language, the first three rows of $\vec{y}_{ls}$ set the left ankle velocity measurement to zero, while the fourth row sets the left ankle $z$ position measurement to floor level, $z_{f}$.
            Note that $\mat{H}_{rs}$ and $\vec{y}_{rs}$ can be constructed in a similar fashion to Eq. \eqref{eq:H-y-lstep-k}.
        \begin{gather}
            \nonumber \\[-0.5em]
            \mat{H}_{mp} = \left[ \begin{array}{crrrc}
                    \bovermat{pelvis, left and right ankle pos. columns}{ \multirow{2}{*}{\phantom{l}$\mat{I}_{3 \times 3}$} & -\tfrac{1}{2}\mat{I}_{2 \times 2} & 0 & -\tfrac{1}{2}\mat{I}_{2 \times 2}}
                     & ... \\
                     & \mat{0}_{1 \times 2} & 0 & \mat{0}_{1 \times 2} & ... 
                \end{array} \right] \label{eq:H-pelv-k} \\
            \vec{y}_{mp} = \begin{bmatrix}
                    \mat{0}_{1 \times 2} & z_{p}
                \end{bmatrix}^T \label{eq:y-pelv-k} \\[1.5em]
            \mat{H}_{ls} = 
                \begin{bmatrix}
                    ...  \bovermat{left ankle pos. and vel. columns}{& \mat{0}_{3 \times 2} & \mat{0}_{3 \times 1} & \mat{I}_{3 \times 3}} & ... \\
                    ... & \mat{0}_{1 \times 2} & 1 & \mat{0}_{1 \times 3} & ... 
                \end{bmatrix}, \quad
            \vec{y}_{ls} = \begin{bmatrix} \mat{0}_{3 \times 1} \\ z_{f} 
                \end{bmatrix} \label{eq:H-y-lstep-k} 
            % \mat{R}_{ls} = \diag(\bv{}{}{\sigma}{2}{ls}) 
        \end{gather}

         As floor contact (FC) varies with time, so too $\mat{H}_{k}$ varies with time, as shown in Eq. \eqref{eq:H-k-cases}.
            Measurements $\vec{y}_{k}$ and state variances  $\bv{}{}{\sigma}{2}{k}$ are constructed similarly to Eq. \eqref{eq:H-k-cases}.
            % \begin{empheq}[box=\myyellowbox]{align}
            \begin{align}
                \mat{H}_{k} = \begin{cases}
                    [ \mat{H}_{mp}^T ]^T & \text{ no FC} \\
                    [ \mat{H}_{mp}^T \quad \mat{H}_{ls}^T ]^T & \text{ left FC} \\
                    [ \mat{H}_{mp}^T \quad \mat{H}_{rs}^T ]^T & \text{ right FC} \\
                    [ \mat{H}_{mp}^T \quad \mat{H}_{ls}^T \quad \mat{H}_{rs}^T ]^T & \text{ both FC} \\
                \end{cases} \label{eq:H-k-cases}
            \end{align}

        The \textit{a priori} state $\kfsm{x}{k}$ is calculated following the traditional KF equations:
        \begin{gather}
            \mat{R}_{k} = \diag(\vec{\sigma}_{k}), \:\:\:
            \mat{K}_{k} = \kfcp{P}{k} \mat{H}_{k}^T \left( \mat{H}_{k} \kfcp{P}{k} \mat{H}_{k}^T + \mat{R} \right)^{-1} \\
            \kfsm{x}{k} = \kfsp{x}{k} + \mat{K}_{k} \left( \vec{y}_{k} - \mat{H}_{k} \kfsp{x}{k} \right)
        \end{gather}
        
        Lastly, the covariance limiter prevents the covariance from growing indefinitely and from becoming badly conditioned, as will happen naturally with the KF tracking the global position of the pelvis and ankles without any global position reference.
            At this step, a pseudo-measurement equal to the current state $\kfsm{x}{k}$ is used (\ie{} $\vec{y}_{lim} = \kfsm{x}{pos,k}$, implemented by $\mat{H}_{lim}$ as shown in Eq. \eqref{eq:H-lim})  with some measurement noise of variance $\bv{}{}{\sigma}{2}{lim}$ ($9 \times 1$ vector).
            The covariance $\kfcm{P}{k}$ is then calculated through Eqs. \eqref{eq:Hkprime}-\eqref{eq:Ptildekprime}.
        \begin{gather}
            \mat{H}_{lim} = \begin{bmatrix} \mat{I}_{9 \times 9} & \mat{0}_{9 \times 9} \end{bmatrix} \label{eq:H-lim} \\
            \mat{H}_{k}' = \begin{bmatrix} \mat{H}_{k}^T & \mat{H}_{lim}^T \end{bmatrix}^T, \quad
            \mat{R}_{k}' = \diag([ \bv{}{}{\sigma}{2}{k} \: \bv{}{}{\sigma}{2}{lim} ]) \label{eq:Hkprime} \\
            \mat{K}_{k}' = \kfcp{P}{k} \mat{H}_{k}'^T \left( \mat{H}_{k}' \kfcp{P}{k} \mat{H}_{k}'^T + \mat{R}' \right)^{-1} \label{eq:Kkprime} \\
            \kfcm{P}{k} = \left(\mat{I} - \mat{K}_{k}' \mat{H}_{k}' \right) \kfcp{P}{k} \label{eq:Ptildekprime}
        \end{gather}
        
    \subsubsection{Satisfying biomechanical constraints} \label{sec:const-nonlin-update}
        After the prediction and measurement updates of the KF, above, the body joints may have become dislocated, or joint angles extend beyond their allowed range. This part of the algorithm corrects the kinematic state estimates to satisfy the biomechanical constraints of the human body. This amounts to projecting the current \textit{a posteriori} state $\kfsm{x}{k}$ estimate onto the constraint surface, guided by our uncertainty in each state variable encoded by $\kfcm{P}{k}$.   
        The constraint equations enforce the following biomechanical limitations: 
        (i) the length of estimated thigh vectors ($||\bv{W}{lt}{\tau}{}{}||$ and $||\bv{W}{rt}{\tau}{}{}||$) equal the thigh lengths $d^{\cs{lt}}$ and $d^{\cs{rt}}$;
        (ii) both knees act as hinge joints; 
        and (iii) the knee joint angle is confined to realistic range of motion and is only allowed to decrease (\ie{} prohibit bending the knee further) to prevent the pelvis drifting towards the floor.
        For the sake of brevity, only the left leg formulation is shown; the right leg formulation can be derived in a similar manner.
    
        Firstly, the constraint for the length of the estimated thigh vector is shown in Eq. \eqref{eq:c-lthigh} where $\bv{W}{lt}{\tau}{}{z}(\kfsc{x}{k})$ is the thigh vector.
            % (a function of the estimated state variables, $\kfsc{x}{k}$).
            The thigh vector (Eq. \eqref{eq:thigh-vect}) was obtained by subtracting the knee joint position from hip joint position.
            % Specifically, it states that the magnitude of thigh vector $||\bv{W}{lt}{\tau}{}{z}(\kfsc{x}{k})||$ subtracted from the expected thigh length equals zero.
            The hip joint position, $\bv{W}{lh}{p}{}{}$, is obtained by adding half the pelvis length along the pelvis $y$ axis, $\bv{W}{p}{r}{}{y}$, from the mid pelvis position, $\bv{W}{mp}{p}{}{}$,
            while the knee joint position, $\bv{W}{lk}{p}{}{}$ is obtained by adding the shank length along its $z$ axis, $\bv{W}{ls}{r}{}{z}$, starting from the ankle joint position, $\bv{W}{la}{p}{}{}$.
            % Refer to Fig. $\ref{fig:body-skeleton-sparse-ckf}$ for visualization.
            As this equation is non-linear, it must be linearized to fit in our KF estimator.
            Linearization is done via Taylor series approximation, as described in \cite[][Sec. 3 Eqs. (66) and (67)]{Simon2010} leading to a system of linear equations $\mat{D}_{ltl, k} \kfsc{x}{k} = d_{ltl, k}$ (Eq. \eqref{eq:c-lthigh-prime} and \eqref{eq:d-lthigh}).
        % \begin{empheq}[box=\myyellowbox]{gather}
        \begin{gather}
            \bv{W}{lt}{\tau}{}{z}(\kfsc{x}{k}) = 
                \overbrace{\bv{W}{mp}{p}{}{} + \tfrac{d^{\cs{p}}}{2} \bv{W}{p}{r}{}{y}}^{\text{hip joint pos.}} - 
                \overbrace{(\bv{W}{la}{p}{}{} + d^{\cs{ls}} \bv{W}{ls}{r}{}{z})}^{\text{knee joint pos.}} \label{eq:thigh-vect}\\
            c_{ltl}(\kfsc{x}{k}) = \sqrt{\bv{W}{lt}{\tau}{}{z}(\kfsc{x}{k})^T \bv{W}{lt}{\tau}{}{z}(\kfsc{x}{k})} - d^{\cs{lt}} = 0 \label{eq:c-lthigh} \\
            \mat{D}_{ltl, k} = \tfrac{\partial c_{ltl}(\kfsc{x}{k})}{\partial \vec{x}}
                = \left[
                    \tfrac{\bv{W}{lt}{\tau}{}{z}(\kfsc{x}{k})}{||\bv{W}{lt}{\tau}{}{z}||} \: 
                    \tfrac{-\bv{W}{lt}{\tau}{}{z}(\kfsc{x}{k})}{||\bv{W}{lt}{\tau}{}{z}||} \: \mat{0}_{1 \times 12} 
                \right] \label{eq:c-lthigh-prime} \\
            d_{ltl, k} = -c_{ltl}(\kfsc{x}{k}) + \tfrac{\partial c_{ltl}(\kfsc{x}{k})}{\partial \vec{x}} \kfsc{x}{k} \label{eq:d-lthigh}
        \end{gather}
        
        Secondly, the constraint for the hinge knee joint enforces the long ($z$) axis of the thigh to be perpendicular to the mediolateral axis ($y$) of the shank, as shown in Eq. \eqref{eq:c-lkhinge}.
            This formulation is similar to \cite[][Sec. 2.3 Eqs. (4)]{Meng2012}.
            Using Eq. \eqref{eq:thigh-vect} to expand the thigh vector $\bv{W}{lt}{\tau}{}{z}$ in Eq. \eqref{eq:c-lkhinge} and rearranging it in terms of $\mat{D}_{lkh, k} \kfsc{x}{k} = d_{lkh, k}$ gives us Eqs. \eqref{eq:D-lkhinge} and \eqref{eq:d-lkhinge}.
            Since the constraint is linear, no linearization is needed.
        % \begin{empheq}[box=\myyellowbox]{gather}
        \begin{gather}
            c_{lkh}(\vec{x}_k) = \bv{W}{lt}{\tau}{}{z} \cdot \bv{W}{ls}{r}{}{y} = 0 \label{eq:c-lkhinge} \\
            \mat{D}_{lkh, k} = \begin{bmatrix}
                \bv{W}{ls}{r}{}{y} & 
                -\bv{W}{ls}{r}{}{y} & \mat{0}_{1 \times 12} 
            \end{bmatrix} \label{eq:D-lkhinge} \\
            d_{lkh, k} = -(\tfrac{d^{\cs{p}}}{2} \bv{W}{p}{r}{}{y}  - d^{\cs{ls}} \bv{W}{ls}{r}{}{z}) \cdot \bv{W}{ls}{r}{}{y} \label{eq:d-lkhinge}
        \end{gather}
        % \bv{W}{mp}{p}{}{} \cdot \bv{W}{ls}{r}{}{y} - \bv{W}{la}{p}{}{} \cdot \bv{W}{ls}{r}{}{y} + (\frac{d^{\cs{p}}}{2} \bv{W}{p}{r}{}{y}  - d_{ls} \bv{W}{ls}{r}{}{z}) \cdot \bv{W}{ls}{r}{}{y}
        
        Thirdly, the constraint for the knee range of motion (ROM) is enforced if the knee angle is outside the allowed ROM or is increasing (\ie{} prohibit bending the knee further).
            This implementation is similar to the active set method used in optimization.
                The knee bending prohibition was implemented to prevent the constraint update state $\kfsc{x}{k}$ from resulting in pose estimates where the pelvis sinks towards the floor.
                Note that the knee can still bend during the prediction and measurement update.
        Mathematically, this is implemented by setting the constrained knee angle $\alpha_{lk}'$ to Eq. \eqref{eq:c-lkrom-cstr} where $\alpha_{lk,min}=0^\circ$ to prevent knee hyperextension and $\alpha_{lk,max}=\mathbf{min}(180^\circ, \hat{\alpha}^+_{lk})$.
            The knee angle $\alpha_{lk}$ is calculated by taking the inverse tangent of the thigh vector, $\bv{W}{lt}{r}{}{z}$, projected on the $z$ and $x$ axes of the shank orientation as shown in Eq. \eqref{eq:c-lkrom-base}.
                It ranges from $-\frac{\pi}{2}$ to $\frac{3\pi}{2}$ as enforced by the chosen sign inside the tangent inverse function and the addition of $\frac{\pi}{2}$. 
            %Fig. \ref{fig:knee-angle-graphic-interp} shows a biomechanical interpretation of $\alpha_{lk}$ from the sagittal plane.
            %\begin{figure}[htbp]
            %    \centering
            %    \includegraphics[width=0.25\textwidth]{figures/body-skeleton-knee-angle.png}
            %    \caption{Knee angle $\alpha_{lk}$ is shown as the angle between the $z$ basis vectors of the thigh, $\bvraw{W}{t}{r}{}{z}$, and shank, $\bvraw{W}{s}{r}{}{z}$, as seen from the plane defined by both basis vectors.}
            %    \label{fig:knee-angle-graphic-interp}
            %\end{figure}
        %If the knee ROM constraint is not satisfied, the knee angle will be set equal to its limit as shown in Eq. \eqref{eq:c-lkrom-limits}.
        %    The lower bound knee angle $\alpha_{lk,min}$ is set to $0^\circ$ to prevent knee hyperextension.
        %    The upper bound knee angle $\alpha_{lk,max}$ is set equal to the knee angle $\hat{\alpha_{lk}}^+$ calculated from the state after the measurement update; or $180^\circ$ if $\hat{\alpha_{lk}}^+$ is greater than $180^\circ$.
        %    \begin{align}
        %        \alpha_{lk}' = \begin{cases}
        %            180^\circ \text{ if } \alpha_{lk} > 180^\circ \\
        %            0^\circ \text{ if } \alpha_{lk} < 0^\circ \\
        %            \hat{\alpha_{lk}}^+ \text{ otherwise}
        %        \end{cases}  \label{eq:c-lkrom-limits}
        %    \end{align}
        Rearranging Eq. \eqref{eq:c-lkrom-base} and setting $\alpha_{lk}=\alpha_{lk}'$ in terms of $\mat{D}_{lkr, k} \kfsc{x}{k} = d_{ltr, k}$ as shown below gives us Eqs. \eqref{eq:D-lkrom} and \eqref{eq:d-lkrom}.
            Note that rearranging Eq. \eqref{eq:c-lkrom-int1} and substituting $\bv{W}{lt}{r}{}{z} = \tfrac{\bv{W}{lt}{\tau}{}{z}(\kfsc{x}{k})}{||\bv{W}{lt}{\tau}{}{z}(\kfsc{x}{k})||}$ gives Eq. \eqref{eq:c-lkrom-int2}.
        % \begin{empheq}[box=\myyellowbox]{gather}
        \begin{gather}
            \alpha_{lk}' = \mathbf{min}(\alpha_{lk,max}, \mathbf{max}(\alpha_{lk,min}, \alpha_{lk})) \label{eq:c-lkrom-cstr} \\
            \alpha_{lk} = \tan^{-1} \left( 
                    \tfrac{ -\bv{W}{lt}{r}{}{z} \cdot \bv{W}{ls}{r}{}{z} }{ -\bv{W}{lt}{r}{}{z} \cdot \bv{W}{ls}{r}{}{x} } 
                \right) + \tfrac{\pi}{2} \label{eq:c-lkrom-base} \\
            \tfrac{ -\bv{W}{lt}{r}{}{z} \cdot \bv{W}{ls}{r}{}{z} }{ -\bv{W}{lt}{r}{}{z} \cdot \bv{W}{ls}{r}{}{x} } = \tfrac{\sin(\alpha_{lk}' - \tfrac{\pi}{2})}{\cos(\alpha_{lk}' - \tfrac{\pi}{2})} \label{eq:c-lkrom-int1} \\
            \bv{W}{lt}{\tau}{}{z} \cdot \overbrace{(\bv{W}{ls}{r}{}{z} \cos(\alpha_{lk}' - \tfrac{\pi}{2}) - \bv{W}{ls}{r}{}{x} \sin(\alpha_{lk}' - \tfrac{\pi}{2}))}^{\text{Denoted as } \vec{\psi}} = 0 \label{eq:c-lkrom-int2} \\
            \mat{D}_{lkr, k} = \begin{bmatrix}
                    \vec{\psi} & -\vec{\psi} & \mat{0}_{1 \times 12} 
                \end{bmatrix} \label{eq:D-lkrom} \\
                d_{lkr, k} = \vec{\psi} \cdot (-\tfrac{d^{\cs{p}}}{2} \bv{W}{p}{r}{}{y} + d^{\cs{ls}} \bv{W}{ls}{r}{}{z}) \label{eq:d-lkrom}
        \end{gather}

        In summary, the linearized constraint equation $\mat{D}_{k} \kfsc{x}{k} = \vec{d}_{k}$ has $\mat{D}_{k} = \begin{bmatrix} \mat{D}_{L, k}^T & \mat{D}_{R, k}^T \end{bmatrix}^T$ and $\vec{d}_{k} = \begin{bmatrix} \vec{d}_{L, k}^T & \vec{d}_{R, k}^T \end{bmatrix}^T$. $\mat{D}_{L, k}$ is shown in Eq. \eqref{eq:D-left} where a bounded knee angle, $\alpha_{lk}$, is defined as $\alpha_{lk,min} < \alpha_{lk} < \alpha_{lk,max}$. 
        $\mat{D}_{R, k}$, $\vec{d}_{L, k}$ and $\vec{d}_{R, k}$ can be derived similarly.
        \begin{align}
            \mat{D}_{L, k} &= \begin{cases}
                [ \mat{D}_{ltl,k}^T \quad \mat{D}_{lkh,k}^T ]^T 
                    & \text{bounded } \alpha_{lk}, \\
                [ \mat{D}_{ltl,k}^T \quad \mat{D}_{lkh,k}^T \quad \mat{D}_{lkr,k}^T ]^T & \text{otherwise.}
            \end{cases} \label{eq:D-left}
        \end{align}
        
        The non-linear constraints for both legs as defined in $\mat{D}_{k} \kfsc{x}{k} = \vec{d}_{k}$ are then enforced through an iterative projection scheme based on the Smoothly Constrained KF (SCKF) \cite{DeGeeter1997}. 
            SCKF applies the standard KF measurement update procedure iteratively, starting with a constraint covariance, $\kfcc{P}{k}$, equal to the \textit{a posteriori} covariance, $\kfcm{P}{k}$.
            The iteration stops when the termination criteria for all constraints are satisfied.
            The termination criterion for each constraint is shown in Eq. \eqref{eq:sckf-threshold} where
                $D_{i,j,k}$ and $\tilde{P}^+_{j,j,k}$ denotes the scalar value at the $i^{\text{th}}$ row and $j^{\text{th}}$ column of $\mat{D}_{k}$ and $\kfcc{P}{k}$, respectively; and
                $\mat{D}_{i,k}$ denotes the $i^{\text{th}}$ row of $\mat{D}_{k}$.
            Please refer to \cite{DeGeeter1997} for a detailed description of SCKF. 
            Note that SCKF can be replaced with other constrained optimization algorithms 
            % (\eg{} Mathworks' Optimization Toolbox's FMINCON function \cite{MatlabFmincon2018}).
            ((\eg{} FMINCON \cite{MatlabFmincon2018}).
            The goal is to satisfy the biomechanical constraints before the next time step, by whatever means.
            \begin{align}
                \tfrac{\mathbf{max}(D_{i,j,k} \tilde{P}^+_{j,j,k} D_{i,j,k})}{\mat{D}_{i,k} \kfcc{P}{k} \mat{D}_{i,k}^T} \geq \bs{}{}{s}{}{thresh,k} \label{eq:sckf-threshold}
            \end{align}
        
    % Note that one can use other estimator machinery.
	\subsection{Post-processing} \label{sec:state-to-ori}
        % The goal of the algorithm is to estimate the orientation of the pelvis, thigh, and shank. 
        The orientation of the pelvis and shank are obtained from the state $\kfsc{x}{k}$. 
        %Since sensors are attached to the pelvis and shank, the orientation of the said segments are the orientation estimation output. 
        The orientation of the left thigh, $\bv{W}{lt}{R}{}{}$, can be calculated using 
        $\bv{W}{lt}{R}{}{} = \begin{bmatrix}
                \bv{W}{ls}{r}{}{y} \times \bv{W}{lt}{r}{}{z} & \bv{W}{ls}{r}{}{y} & \bv{W}{lt}{r}{}{z}
            \end{bmatrix}$ 
        where $\bv{W}{lt}{r}{}{z} = \tfrac{\bv{W}{lt}{\tau}{}{}}{||\bv{W}{lt}{\tau}{}{}||}$. 
        The orientation of the right thigh, $\bv{W}{rt}{R}{}{}$, is calculated similarly.

	\section{Experiment}
	    The goal of the experiment is to evaluate the algorithm described in this paper against benchmark systems, focusing only on the kinematics of the pelvis, thighs, and shanks.
	    
	    \subsection{Setup}
\begin{figure}[htbp]
	            \centering
	            \includegraphics[height=4cm]{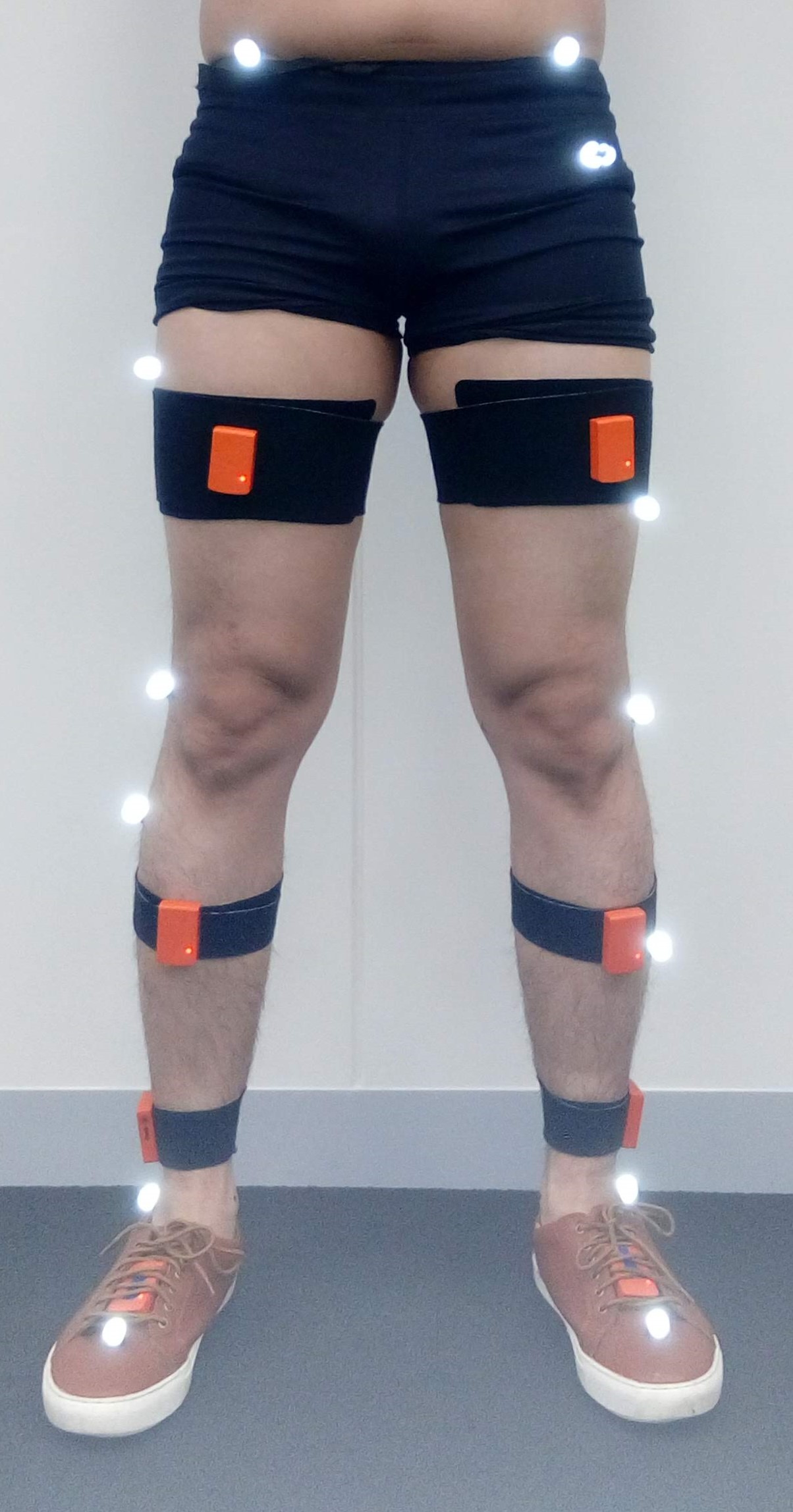}
	            \includegraphics[height=4cm]{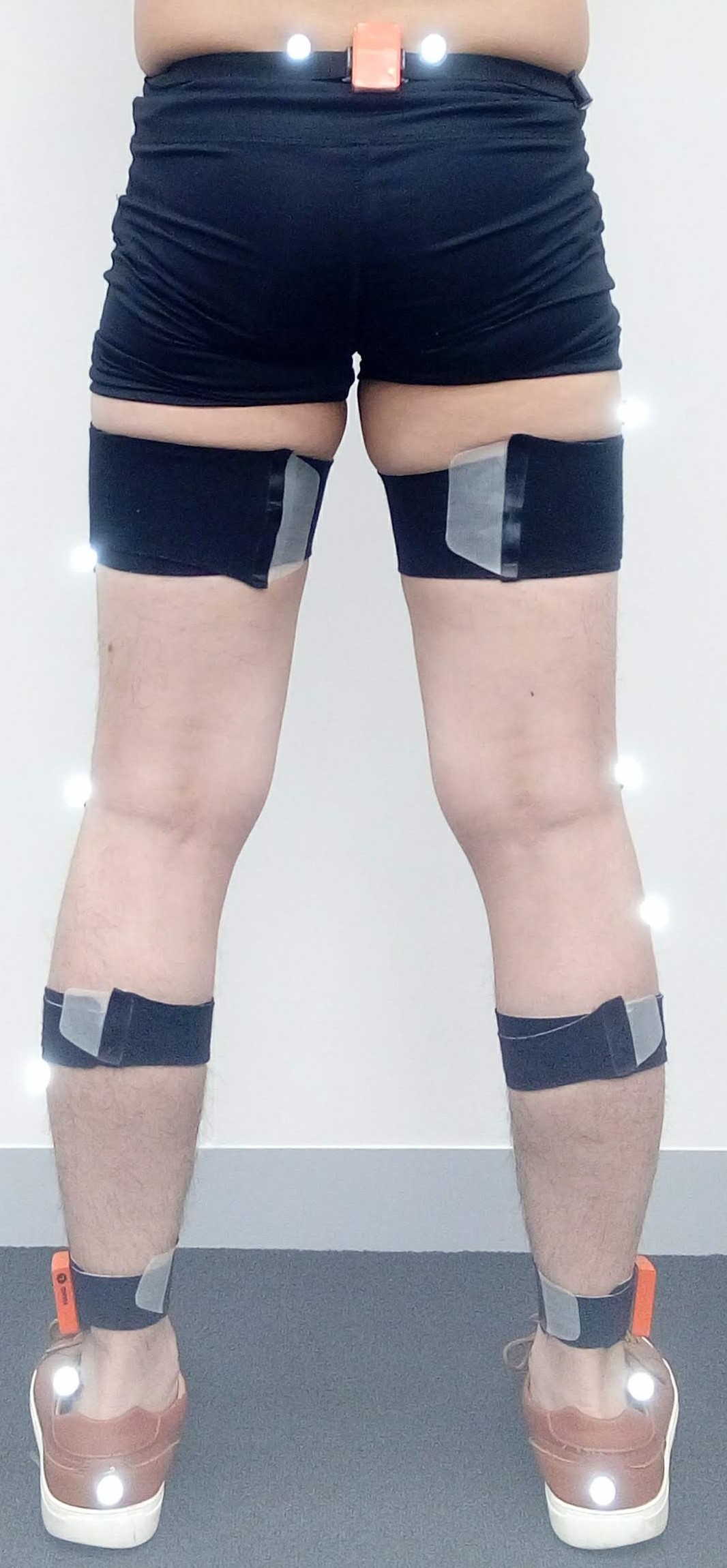}
	            \caption{Sample front and back view of the \changed{OMC (\ie{} Vicon)} reflective markers and the \changed{OSPC (\ie{} Xsens)} IMU setup used in validation experiments.}
	            \label{fig:experiment-setup}
	        \end{figure}

	        %The experiment had $10$ subjects ($7$ males and $3$ females with $60.97 \pm 8.9$ kg weight and $1.68 \pm 0.08$ m height).
	        The experiment had nine healthy subjects ($7$ men and $2$ women, weight $63.0 \pm 6.8$ kg, height $1.70 \pm 0.06$ m, age $24.6 \pm 3.9$ years old), with no known gait or lower body biomechanical abnormalities.
	        Our system was compared to two benchmark systems, namely \changed{an OMC (\ie{} Vicon) and OSPS (\ie{} Xsens)} systems. 
	            The Vicon Vantage system consisted of eight cameras covering approximately $4 \times 4$ m$^2$ capture area with millimetre accuracy. 
	                Vicon data were captured at $100$ Hz and processed using Nexus 2.7 software.
	            The Xsens Awinda system consisted of seven MTx units (IMUs). 
	                Xsens data were captured at $100$ Hz using MT Manager 4.8 and processed using MVN Studio 4.4 software.
	            % with accuracy according to specification.
	            The Vicon and Xsens recordings were synchronized by having the Xsens Awinda station send a trigger pulse to the Vicon system at the start and stop event of each recording \cite{SyncXsenswithVicon2015}.
            Each subject had reflective Vicon markers placed according to the Helen-Hayes 16 marker set \cite{Kadaba1990}, 
                seven MTx units attached to the pelvis, thighs, shanks, and feet according to standard Xsens sensor placement, 
                and two MTx units attached near the ankles, as shown in Fig. \ref{fig:experiment-setup}.
                The MTx units were all factory calibrated. 
                % Acceleration and orientation where obtained using the MTx units and the black-box algorithm that came with it. 
            Note that there are many published works that have validated the Xsens system against OMC systems \cite{Cloete2008, VanDenNoort2013}.
	        
	        Each subject performed the movements listed in Table \ref{tab:movement-type-desc} twice (\ie{} two trials). 
	            The subjects stood still before and after each trial for ten seconds.
	            The experiment was approved by the Human Research Ethics Board of the University of New South Wales (UNSW) with approval number HC180413.
	        \begin{table}[htbp]
	            \centering
	            \caption{Types of movements done in the validation experiment}
	            \label{tab:movement-type-desc}
	            \begin{tabular}{cll}
	                \hline \hline
	                 Movement & Description & Duration (s) \\ \hline
	                 Static & Stand still & $\sim 10$ \\ %& N \\
	                 Walk & Walk straight and back & $\sim 30$ \\%& N \\
	                 Figure of eight & Walk in figures of eight & $\sim 60$ \\%& N \\
	                 Zig-zag & Walk zigzag & $\sim 60$ \\%& N \\
	                 5-minute walk & Undirected walk, side step, and stand & $\sim 300$ \\%& N \\
	                 %Jog straight and back & $\sim 30$ & Y \\
	                 %High knee jogging & $\sim 30$ & Y \\
	                 %Jumping jacks & $\sim 30$ & Y \\
	                 %Speed skater & $\sim 30$ & Y \\
	                 \hline \hline
	            \end{tabular}
	        \end{table}
            % static; walk and jog straight and back to starting point; walk in figure of eight and zigzag around $4 \times 4$ meter field; jumping jacks, speed skater, and high knee jogging in place; free walking for five minutes. Each trial (except the free walking) took around 30 seconds of motion.
	        
	    \subsection{Calibration}
	        \subsubsection{Frame alignment}  \label{sec:calib-frame-align}
	        A common reference frame is needed for objective pose comparison.
	        In this experiment, comparison was done in the world frame, $\cs{W}$, 
	            which was set equal to the Xsens frame.
	        The pose estimate from the \changed{OMC} system which was in the vicon frame, $\cs{V}$, was converted to the world frame, $\cs{W}$, using a fixed rotational offset, $\quat{W}{V}{}{0}$,
	            which was calculated from a calibration procedure where the direction of gravity and the Earth's magnetic field were measured using a compass and pendulum with optical markers attached \cite{madgwick2010efficient}.

            \subsubsection{Yaw offset} \label{sec:calib-yaw-offset}
            Due to varying magnetic interference between the ankle and pelvis, an additional calibration procedure was executed to ensure that the $x$ axes (north) of all sensors were aligned. 
            The sensors in the ankle were especially affected, with yaw offset (\ie{} rotation offset around the $z$ axes) errors due to this magnetic disturbance.
            A grid search was performed to find the yaw offset that produces the least $x$ and $y$ axis RMSE between the \changed{OMC} inferred acceleration and the IMU acceleration measurement after adjusting for each candidate yaw offset (as shown in Eq. \eqref{eq:yaw-offset-rmse} for the left ankle 
            where $\mathring{\vec{\tau}}=\begin{bmatrix} 0 & \vec{\tau}^T \end{bmatrix}^T$,
                        \mbox{$\qmult$} and \mbox{\makebox[\widthof{q}]{}$^{-1}$} are the quaternion and inversion operator, respectively).
            The optimal yaw offset was then used to adjust the corresponding shank orientations.
            Note that the calibration was done against the \changed{OMC} benchmark system to minimize pose differences due to calibration and ensure an objective evaluation of the CKF algorithm.
            \begin{align}
                \mathbf{RMSE}_{x,y} ( \bv{W}{la}{\mathring{a}}{}{\rm{omc}} - ( \quat{}{}{}{\rm{yaw}} \qmult \bv{W}{la}{\mathring{\breve{a}}}{}{\rm{imu}} \qmult \quat{}{}{-1}{\rm{yaw}} ) ) \label{eq:yaw-offset-rmse}
            \end{align}
            
        \subsection{System parameters}
            The proposed algorithm used measurements from the three MTx units attached to the pelvis and ankles. The algorithm and calculations were implemented using Matlab 2018b.
            
            \subsubsection{Body segment orientation}
                Since our goal is to evaluate the tracking algorithm, not the initialization procedure, the initial pose of the body segments, $\quat{W }{B}{}{0}$, was obtained from the \changed{OMC} system where that subject was asked to stand in an N pose (standing upright, with hands by sides).
                    % see Eq. \eqref{eq:body-ori-in-world-frame} on how $\quat{W }{B}{}{0}$ was used to calculate the orientation of the body segment.
                The measured orientations, $\mathbb{IO}$, were obtained using the orientation \changed{estimates} provided by the Xsens system \cite{Roetenberg2009}.
                
            \subsubsection{Step detection}
                The algorithm used for this experiment detects a step if the variance of the free body acceleration across the $0.25$ second window is below a threshold. The threshold was set to $1$ m.s$^{-2}$.
                The step detection sometimes detect steps even if the foot/ankle was moving. 
                    To prevent such false positives, the step detection output was manually reviewed and corrected.
                
            \subsubsection{State estimator}
                The position and velocity of $\kfsc{x}{0}$ were set to the initial position and velocity obtained from the \changed{OMC} benchmark system.
                % while the orientation was set to zero (\ie{} $\quat{W}{i}{}{0} = \vec{0}_{4 \times 1}$).
                $\kfcm{P}{0}$ was set to $0.5 \mat{I}_{30 \times 30}$.
                The SCKF threshold, $\bs{}{}{s}{}{thresh,k}$, was set to $100$.
                The variance parameters used to generate the process and measurement error covariance matrix $\mat{Q}$ and $\mat{R}$ are shown in Table \ref{tab:var-param-kfnoise}.
                \begin{table}[htbp]
                    \centering
                    \caption{Variance parameters for generating the process and measurement error covariance matrices, $\mat{Q}$ and $\mat{R}$. }
                    \begin{tabular}{c|ccc} \hline \hline
                        \multicolumn{1}{c|}{$\mat{Q}$ Parameters} &
                        \multicolumn{3}{c}{$\mat{R}$ Parameters}
                        \\ \hline
                        $\bv{}{}{\sigma}{2}{acc}$ & 
                        $\bv{}{}{\sigma}{2}{mp}$ &
                        $\bv{}{}{\sigma}{2}{ls}$ and $\bv{}{}{\sigma}{2}{rs}$ &
                        $\bv{}{}{\sigma}{2}{lim}$
                        \\
                        (m$^2$.s$^{-4}$) & 
                        (m$^2$) &
                        ([m$^2$.s$^{-2}$ m$^2$]) &
                        (m$^2)$
                        \\ \hline
                        $10^2\mat{1}_{9}$ & 
                        $[10^2 \: 10^2 \: 0.1]$ & 
                        $[0.01\mat{1}_{3} \: 10^{-4}]$ & 
                        $10^2 \mat{1}_{9}$ \rule{0pt}{2.6ex} \\
                        \hline \hline
                    \end{tabular} 
                    
                    where $\mat{1}_{n}$ is an $1 \times n$ row vector with all elements equal to $1$
                    \label{tab:var-param-kfnoise}
                \end{table}

        \subsection{Evaluation Metrics}
            In order to compare the algorithm's accuracy against an existing motion capture system that uses fewer sensors, the following metrics were chosen. 
            \subsubsection{Mean position RMSE}
                is a common metric in video based human motion capture systems (\eg{} \cite{Marcard2017}), commonly denoted as $d_{pos}$.
                In this paper, it is denoted as $e_{pos}$ to prevent confusion with body segment length symbols.
                However, as our system only captures the lower body, the metric has been modified as shown in Eq. (\ref{eq:d-pos}) 
                    where $\mathbb{DP}$ = $\{ lh, rh, lk, rk, la, ra \}$,
                    $\bv{W}{i}{p}{}{k}$ is the benchmark position, 
                    and $\bv{W}{i}{\tilde{p}}{}{k}$ is the estimated position.
                Six body points, $N_{pos} = 6$, were evaluated instead of thirteen body points (full body).
                Furthermore, as the global position of the estimate is still prone to drift due to the absence of an external global position reference, the root position of our system was set equal to that of the benchmark system (\ie{} the  mid-pelvis is placed at the origin for all RMSE calculations).
                \begin{align}
                    e_{pos, k} = \tfrac{1}{N_{pos}} \textstyle \sum_{i \in \mathbb{DP}} ||\bv{W}{i}{p}{}{k} - \bv{W}{i}{\tilde{p}}{}{k} || \label{eq:d-pos}
                \end{align}
                
            \subsubsection{Mean orientation RMSE}
                $e_{ori}$ are the exponential coordinates of the rotational offset between the estimated and benchmark orientation of the body segments which do not have IMUs attached to them (\ie{}  the thighs, $\mathbb{DO}$ = $\{ \quat{W}{lt}{}{k} \quat{W}{rt}{}{k} \}$ and $N_{ori} = 2$). 
                This metric was calculated similarly to \cite{Marcard2017}, as shown in Eq. \eqref{eq:d-ori}, 
                    where $\quat{W}{i}{}{k}$ is the benchmark orientation,
                    $\quattilde{W}{i}{}{k}$ is the estimated orientation,
                    and $\text{v}$-operator extracts the coordinates of the skew-symmetric matrix obtained from the matrix logarithms operation.
                In practical implementation, the quaternion inside the logarithm function is converted to rotation matrix representation.
                \begin{align}
                    e_{ori, k} = \tfrac{1}{N_{ori}} \textstyle \sum_{i \in \mathbb{DO}} || \mathbf{log} \left( \quat{W}{i}{}{k} \qmult \quattilde{W}{i}{-1}{k} \right)^{\text{v}} || \label{eq:d-ori}
                \end{align}
                
            \subsubsection{Hip and knee joint angles RMSE and coefficient of correlation (CC)}
                The hip joint angles in the sagittal, frontal, and transverse plane and knee joint angle in the sagittal plane are commonly used parameters in gait analysis. 
                Hence, the angle RMSE and CC are useful indicators of the accuracy of the system in clinical applications. 
                CC of joint $i$ is obtained using Eq. \eqref{eq:corrcoef}, 
                    where $\theta^{i}_{k}$ and $\tilde{\theta}^{i}_{k}$ are the benchmark angle
                    and the estimated angle of joint $i$ at time step $k$, respectively.
                \begin{align}
                    \text{CC}^{i} = \tfrac{n \sum \theta^{i}_{k} \tilde{\theta}^{i}_{k} 
                        - \left(\sum \theta^{i}_{k} \right)
                          \left(\sum \tilde{\theta}^{i}_{k} \right)}
                    {\sqrt{n \left( \sum (\theta^{i}_{k})^2 \right) 
                            - \left( \sum \theta^{i}_{k} \right)^2} 
                     \sqrt{n \left(\sum (\tilde{\theta}^{i}_{k})^2 \right) 
                            - \left( \sum \tilde{\theta}^{i}_{k} \right)^2}} \label{eq:corrcoef}
                \end{align}
        \subsubsection{Total travelled distance (TTD) deviation}
        \changed{
            is the TTD error with respect to the actual TTD. 
            TTD is calculated by summing the distance between the XY positions of the pelvis and ankles at specific event times. 
                The events for the pelvis and left ankle are when left footsteps are detected, 
                while the events for the right ankle are when right footsteps are detected.
        }
        
    \section{Results}
        \subsection{Mean position and orientation RMSE}
\begin{figure}[htbp]
                % \begin{mdframed}[style=mdchange]
                \centering
                \includegraphics[width=\linewidth]{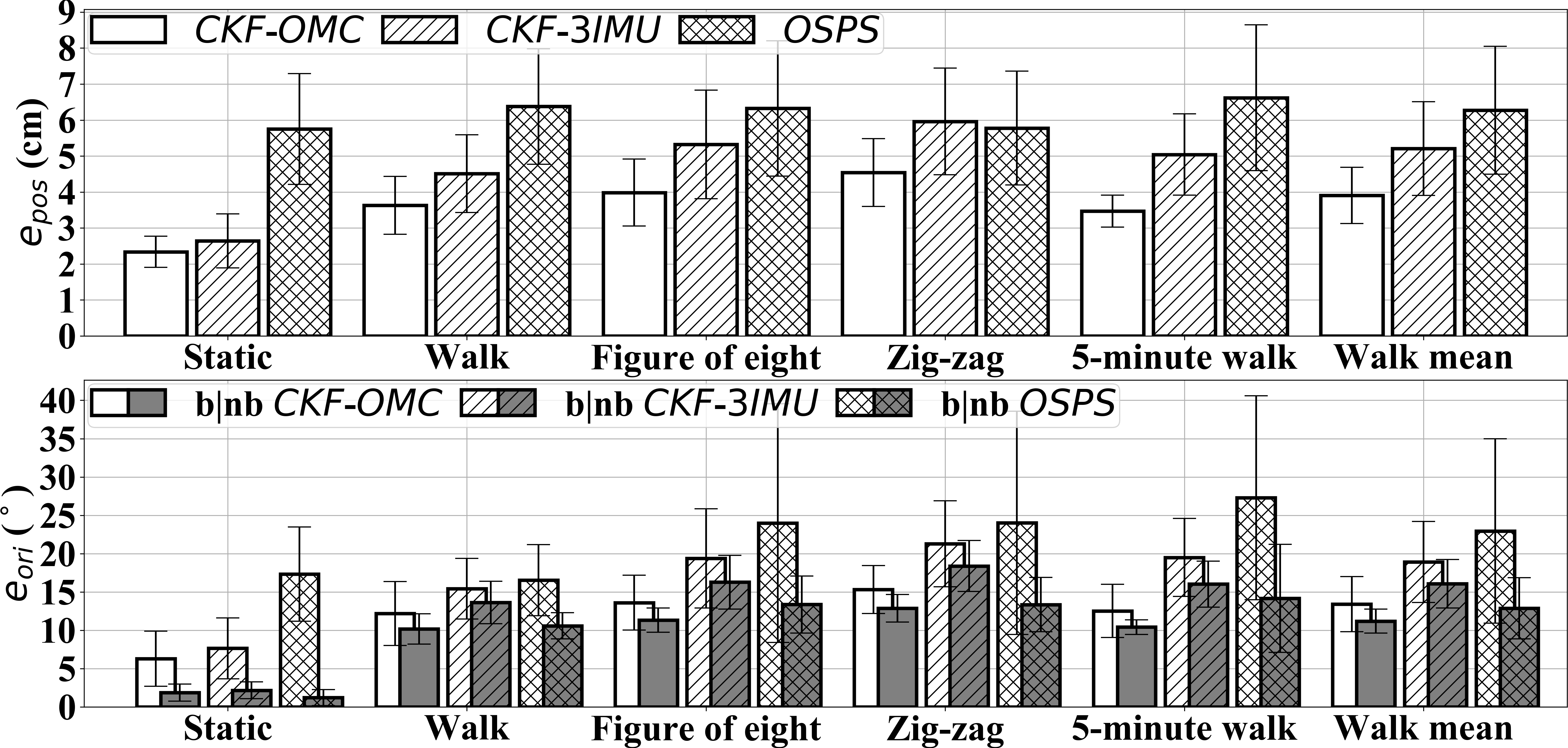}
                \caption{The mean position and orientation RMSE, $e_{pos}$ (top) and $e_{ori}$ (bottom), for \emph{CKF-OMC}, \emph{CKF-3IMU}, and \emph{OSPS} of each motion type. The prefix b denotes biased, while nb denotes no or without bias.}
                % \end{mdframed}
                \label{fig:results-dposdori}
            \end{figure}

            Fig. \ref{fig:results-dposdori} shows the mean position and orientation RMSE of our algorithm for each movement type.
            The comparison involved the output from three configurations of interest, all compared against the \changed{OMC} gold standard output: 
                i) our algorithm using acceleration (double derivative of position) and orientation obtained from \changed{OMC} (denoted as \emph{CKF-OMC});
                ii) our algorithm using raw 3D acceleration in sensor frame and orientation in world frame obtained from Xsens MTx IMU measurements (denoted as \emph{CKF-3IMU}), and;
                iii) the black box output from the MVN Studio software (denoted as \emph{OSPS}).
                \emph{CKF-OMC} gives insight into the best possible output of the CKF algorithm when given acceleration and orientation which corresponds perfectly with the position measured by the \changed{OMC} benchmark system, 
                while the \emph{OSPS} illustrates the performance of a widely accepted commercial wearable HMCS with an OSPS configuration.
            Both biased and unbiased (\ie{} for unbiased, the mean difference between the angles over each entire trial was subtracted) $e_{ori}$ are presented to account for different anatomical calibration offset errors between the \changed{OMC} and \changed{OSPS} systems \cite{Cloete2008, VanDenNoort2013}.
            
        \subsection{Hip and knee joint angle RMSE and CC}
\begin{figure}[htbp]
                % \begin{mdframed}[style=mdchange]
                \centering
                \includegraphics[width=\linewidth]{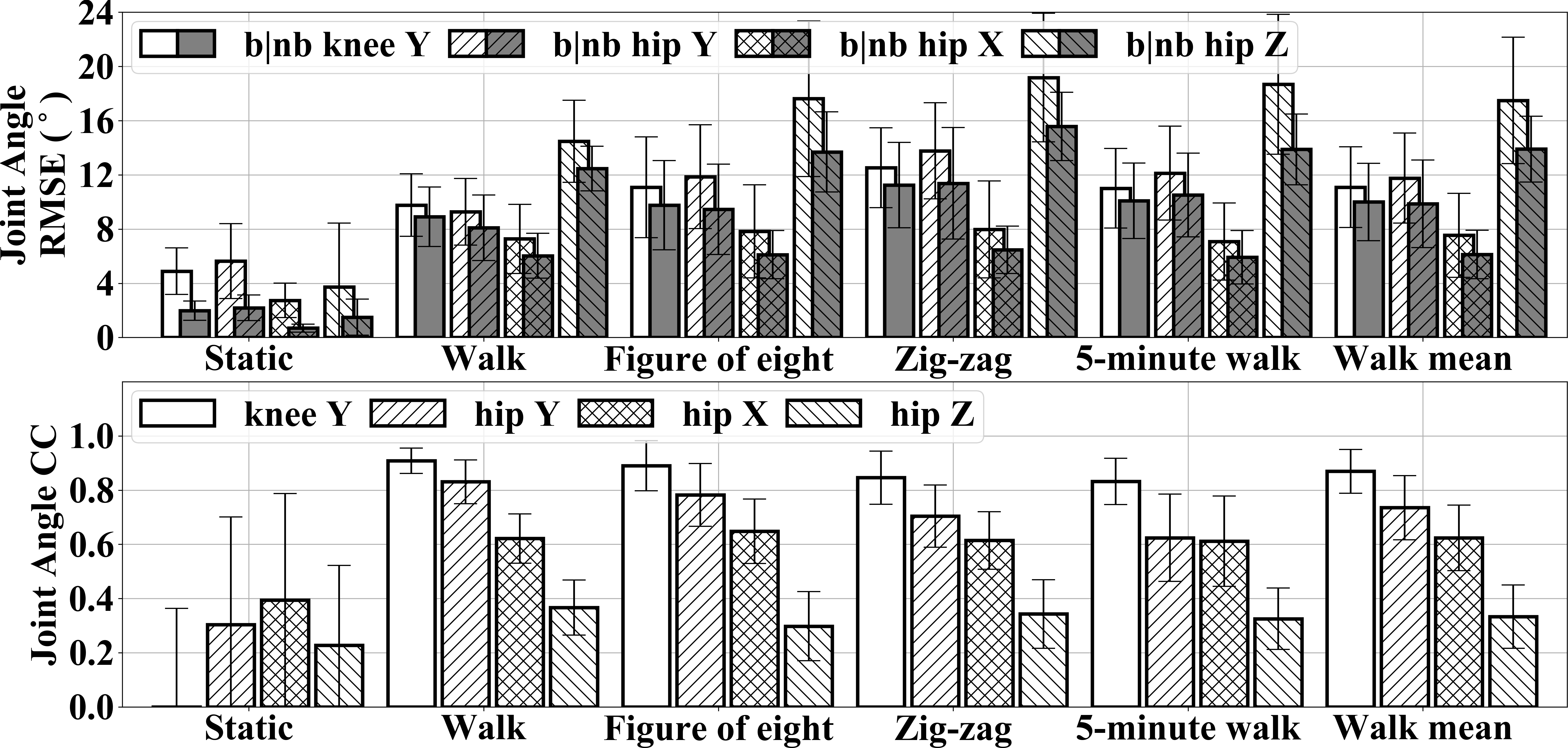}
                \caption{The joint angle RMSE (top) and CC (bottom) of knee and hip joint angles for \emph{CKF-3IMU} at each motion type. Y, X, and Z denotes the sagittal, frontal, and transversal plane, respectively.
                The prefix b denotes biased, while nb denotes no bias.}
                % \end{mdframed}
                \label{fig:results-kneehip-angles-rmsecc}
            \end{figure}

\begin{figure}[htbp]
                % \begin{mdframed}[style=mdchange]
                \centering
                \includegraphics[width=\linewidth]{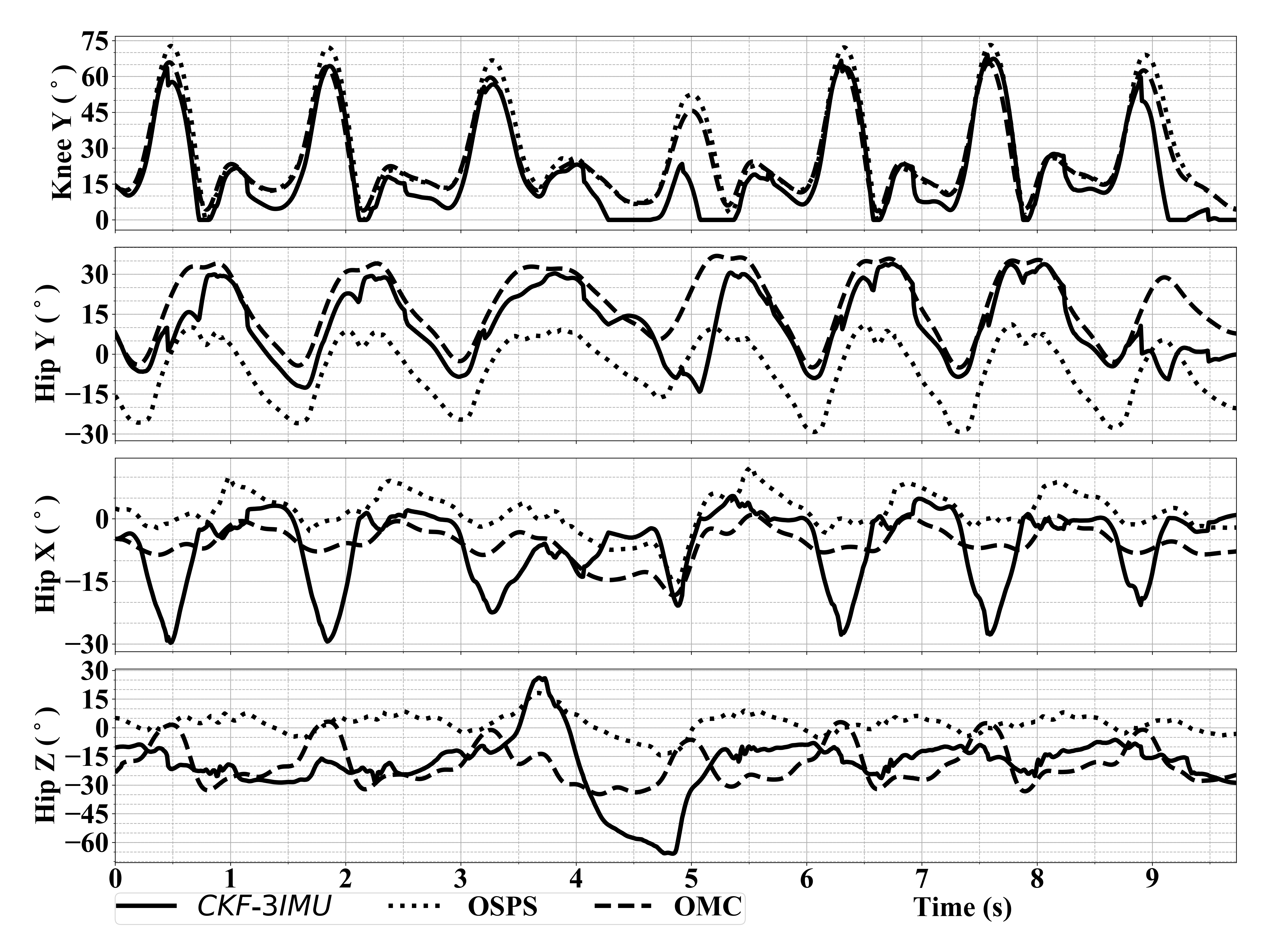}
                \caption{Knee and hip joint angle output of \emph{CKF-3IMU} in comparison with \emph{OSPS} and the benchmark system (\changed{OMC}) for a \emph{Walk} trial. The subject walked straight from $t=0$ to $3$ s, turned $180^\circ$ around from $t=3$ to $5.5$ s, and walked straight to original point from $5.5$ s until the end of the trial.}
                \label{fig:results-kneehip-angle-sample}
                % \end{mdframed}
            \end{figure}

            Fig. \ref{fig:results-kneehip-angles-rmsecc} shows the knee and hip joint angle RMSE and CC for \emph{CKF-3IMU}.
                Y, X, and Z refers to the plane defined by the normal vectors $y$, $x$, and $z$ axes, respectively,
                    and are also known as the sagittal, frontal, and transversal plane in the context of gait analysis.
            Fig. \ref{fig:results-kneehip-angle-sample} shows a sample \emph{Walk} trial.
            \vspace{-6pt}

        \subsection{Total travelled distance (TTD) deviation}
        \changed{
            Table \ref{tab:results-ttddev} shows the TTD deviation for \emph{CKF-3IMU} at the pelvis and ankles along with related literature \cite{Jimenez2010a, Zhang2017a}.
            Refer to the supplementary material for a tvideo reconstruction of sample trials \cite{Sy-ckf2019-supp}.
        }
        
        % Table generated by Excel2LaTeX from sheet 'TTD'
        \begin{table}[htbp]
            \centering
            \caption{Total travelled distance deviation for \emph{CKF-3IMU} at the pelvis and ankles along with related literature}
            % \begin{mdframed}[style=mdchange]
            \begin{tabular}{ccccc}
            \hline
            \hline
            \multicolumn{1}{c}{Pelvis} & \multicolumn{1}{c}{Left Ankle} & \multicolumn{1}{c}{Right Ankle} & \multicolumn{1}{c}{\cite{Jimenez2010a}} & \multicolumn{1}{c}{\cite{Zhang2017a}} \bigstrut\\
            \hline
            4.93\% & 3.81\% & 3.60\% & 0.3-1.5\% & 0.21-0.35\% \bigstrut\\
            \hline
            \hline
            \end{tabular}%
            % \end{mdframed}
            \label{tab:results-ttddev}%
        \end{table}%
	        
	\section{Discussion}
	    In this paper, a CKF algorithm for lower body pose estimation using only three IMUs, ergonomically placed on the ankles and sacrum to facilitate continuous recording outside the laboratory, was described and evaluated. The algorithm utilizes fewer sensors than other approaches reported in the literature, at the cost of reduced accuracy.
        
    % comparison
    \subsection{Mean position and orientation RMSE comparison}
        The mean position and orientation RMSE, $e_{pos}$ and $e_{ori}$, of \emph{CKF-OMC}, \emph{CKF-3IMU}, \emph{OSPS}, and related literature for walking related motions (\ie{}  \emph{Walk}, \emph{Figure of eight}, \emph{Zig-zag}, and \emph{5-minute walk}) are shown in Table \ref{tab:e-pos-e-ori} \cite{Marcard2017}.
            The $e_{pos}$ and $e_{ori}$ (no bias) difference between \emph{CKF-OMC} and \emph{CKF-3IMU} were around $1$ cm and $5^\circ$, respectively, indicating that our estimator was able to handle sensor noise well when taking orientation and acceleration readings from the IMUs, rather than being derived from the \changed{OMC} position measurements.
                Note that the relatively large orientation error for \emph{CKF-OMC} was caused by the constant body segment length and hinge knee joint assumption; measuring body segments as the distance between adjacent joint centre positions, as estimated by the Vicon software, results in segment lengths which vary slightly throughout the gait cycle.
            The \emph{OSPS} errors were comparable to the results of CKF-OMC algorithm. 
                There was significant improvement in the  $e_{ori}$ results of the OSPS when bias was removed from this error variable; 
                this bias probably due to the difference in the sensor-to-body-segment rotational offset, $\quat{B}{S}{}{0}$, 
                caused by OSPS\mbox{’} (\ie{} Xsens\mbox{’}) black-box pre-processing of body segment orientation (Sec. \ref{sec:body-seg-ori}) \cite{Cloete2008, VanDenNoort2013}.
            \emph{CKF-3IMU}'s performance for motion \emph{5-minute walk} in comparison to other shorter duration movements shows its tracking capability even over long durations; 
                note all errors are measured locally, with the mid-pelvis as the origin.
                % This feature is desirable as running and walking gait analysis should involve long-term data collection on large population of subjects, thus enabling the identification of features/characteristics truly indicative of the disorder under interest \cite{Benson2018}.
        
        \emph{CKF-3IMU} was compared to two algorithms from the study of Marcard \etal{} \cite{Marcard2017}; namely, sparse orientation poser (\emph{SOP}) and sparse inertial poser (\emph{SIP}).
            Note that \emph{SOP} used orientation measured by IMUs and biomechanical constraints,
                while \emph{SIP} used similar information but with the addition of acceleration measured by IMUs.
            Both \emph{SOP} and \emph{SIP} were benchmarked against \changed{an OSPS} system while our algorithm was benchmarked against \changed{an OMC} system.
            The $e_{pos}$ and $e_{ori}$ (no bias) difference between \emph{SOP} and \emph{CKF-3IMU} were \texttildelow$0.2$ cm and $1^\circ$, respectively (\ie{} comparable).
                % The better mean orientation RMSE, $e_{ori}$, may be due to \emph{SOP}'s utilization of a less restrictive biomechanical constraint.
            However, the $e_{pos}$ and $e_{ori}$ (no bias) of \emph{SIP} were smaller than \emph{CKF-3IMU} by \texttildelow$2.2$ cm and $5^\circ$, respectively.
                This improvement was expected, as \emph{SIP} optimizes the pose over multiple frames whereas \emph{CKF-3IMU} optimizes the pose for each individual frame.
    	        Comparing processing times, \emph{CKF-3IMU} was dramatically faster than \emph{SIP}; 
    	            \emph{CKF-3IMU} processed a 1,000-frame sequence in less than a second on a Intel Core i5-6500 3.2 GHz CPU, while \emph{SIP} took 7.5 minutes on a quad-core Intel Core i7 3.5 GHz CPU. 
    	            Both set-ups used single-core non-optimized MATLAB code.
	            The \emph{CKF-3IMU} could hence be used as part of a  wearable HMCS which could provide real-time gait parameter measurement to inform actuation of assistive or rehabilitation robotic devices.
	            
        % Table generated by Excel2LaTeX from sheet 'Summary'
        \begin{table}[htbp]
            \centering
            % \begin{mdframed}[style=mdchange]
            \caption{Mean position and orientation RMSE of \emph{CKF-OMC}, \emph{CKF-3IMU}, \emph{OSPS}, Sparse orientation and inertial poser (\emph{SOP} and \emph{SIP}) \cite{Marcard2017}}
            \label{tab:e-pos-e-ori}%
            % Table generated by Excel2LaTeX from sheet 'Summary-v2'
            \begin{tabular}{p{7em}rrrrr}
            \hline
            \hline
            \multicolumn{1}{l}{} & \multicolumn{1}{c}{\emph{CKF-OMC}} & \multicolumn{1}{c}{\emph{CKF-3IMU}} & \multicolumn{1}{c}{\emph{OSPS}} & \multicolumn{1}{c}{\emph{SOP}} & \multicolumn{1}{c}{\emph{SIP}} \bigstrut\\
            \hline
            \multicolumn{1}{l}{$e_{pos}$ (cm)} & $3.9 \pm 0.8$ & $5.2 \pm 1.3$ & $6.3 \pm 1.8$ & \multicolumn{1}{l}{\texttildelow $5.0$} & \multicolumn{1}{l}{\texttildelow $3.0$} \bigstrut\\
            $e_{ori}$ ($^\circ$) no bias & $11.2 \pm 1.6$ & $16.1 \pm 3.2$ & $12.9 \pm 4.0$ & \multicolumn{1}{l}{\texttildelow $15.0$} & \multicolumn{1}{l}{\texttildelow $11.0$} \bigstrut\\
            mean* ($^\circ$) & $0.6 \pm 1.4$ & $1.2 \pm 1.9$ & $-2.5 \pm 6.7$ & -     & - \bigstrut\\
            $e_{ori}$ ($^\circ$) biased & $13.4 \pm 3.6$ & $18.9 \pm 5.3$ & $23.0 \pm 12.0$ & -     & - \bigstrut\\
            \hline
            \hline
            \end{tabular}%
            
            * aggregated on \changed{per-trial} level
            % \end{mdframed}
        \end{table}%

    \subsection{Hip and knee joint angle RMSE and CC comparison}
        The knee and hip joint angle RMSEs and CCs of \emph{CKF-3IMU}, \emph{OSPS}, and related literature are shown in Table \ref{tab:knee-hip-angle-rmse-cc} \cite{Cloete2008, Hu2015, Tadano2013}. 
            The RMSE difference ($<6^\circ$) between \emph{CKF-3IMU} and \emph{OSPS} indicates that our estimator is robust to handle sensor noise.
            The CC of \emph{CKF-3IMU}, specifically the knee and hip joint angles in the sagittal plane ($0.87$ and $0.74$), indicates that these angles have good correlation with the \changed{OMC} benchmark system .
            Similar qualitative observations can be made in Fig. \ref{fig:results-kneehip-angle-sample}.
                The difference between \emph{CKF-3IMU} and the \changed{OMC} benchmark on the frontal and transverse plane were probably due to the strict hinged knee joint assumption, which is not perfectly anatomically accurate; 
                the knee has a moving center \changed{of} rotation more accurately approximated as a rolling contact between the shank and thigh segments. 
                The hinge joint does not capture the moving center of rotation and contributes to estimation error.
                %the knee joints more resembles a four-bar linkage, and can also experience a small amount of medial and lateral. 
                Violations of this assumption are apparent at $t=3$ to $5.5$ s in Fig. \ref{fig:results-kneehip-angle-sample}, where the subject makes a $180^\circ$ turn.
                After the increased angular error due to the $180^\circ$ turn, \emph{CKF-3IMU} was able to recover during the straight walking ($t=5.5$ to $9.74$ s of Fig. \ref{fig:results-kneehip-angle-sample}).
            Notice that the bias between OSPS and \changed{OMC} can be observed in Fig. \ref{fig:results-kneehip-angle-sample} at $t=0$ of the hip angle waveforms. 
                The results also agree with the study of Cloete \etal{}, namely the smaller CC and larger RMSEs (both biased and no bias) for hip joint angles in the frontal and transverse planes \cite{Cloete2008}.
        
        Considering the study of Hu \etal{} and Tadano \etal, the errors were $1^\circ$ to $6^\circ$ smaller than \emph{CKF-3IMU} \cite{Hu2015, Tadano2013}.
            Hu \etal{} used 4 IMUs (two  at  pelvis  and one for each foot) and Tadano \etal{} used an OSPS configuration.
            However, their trials were limited to a single gait cycle, in contrast to the longer-duration walking trials in this paper which had at least five gait cycles involving turns and capture duration of $1$ to $5$ minutes.
	        % Furthermore, these systems can only estimate the pose in the sagittal plane.
	        
        Lastly, it is important to note that although \emph{CKF-3IMU} achieves good joint angle CCs in the sagittal plane, the joint angle RMSE ($>5^\circ$) makes its utility in clinical applications uncertain. 
            To achieve clinical utility, one may leverage long-term recordings by averaging out cycle-to-cycle variation in estimation errors over many gait cycles, 
            or consider only body movements from which kinematics that can be estimated more accurately. 
            The extent to which these possible solutions can bridge the gap to clinical application for the proposed system remains to be seen in future work.
        \begin{table}[htbp]
            % \begin{mdframed}[style=mdchange]
            \centering
            \caption{Knee and hip angle RMSE (top) and CC (bottom) of \emph{CKF-3IMU}, \emph{OSPS}, and related literature}
            % Table generated by Excel2LaTeX from sheet 'Summary-v2'
            \begin{tabular}{lrrrrr}
            \hline
            \hline
            \multicolumn{2}{p{5em}}{Joint Angle RMSE ($^\circ$)} & \multicolumn{1}{c}{knee sagittal} & \multicolumn{1}{c}{hip sagittal} & \multicolumn{1}{c}{hip frontal} & \multicolumn{1}{c}{hip transverse} \bigstrut\\
            \hline
            \multicolumn{1}{p{2.5em}}{\multirow{3}[1]{*}{\parbox{2.5em}{\emph{CKF-}\newline{}\emph{3IMU}}}} & \multicolumn{1}{l}{biased} & $11.1 \pm 3.0$ & $11.8 \pm 3.3$ & $7.5 \pm 3.1$ & $17.5 \pm 4.7$ \bigstrut[t]\\
                  & \multicolumn{1}{l}{mean} & $-1.2 \pm 4.2$ & $-4.3 \pm 4.4$ & $-2.2 \pm 4.2$ & $-4.0 \pm 9.7$ \\
                  & \multicolumn{1}{l}{no bias} & $10.0 \pm 2.9$ & $9.9 \pm 3.2$ & $6.1 \pm 1.8$ & $13.9 \pm 2.4$ \\ \hdashline
            \multirow{3}[0]{*}{\emph{OSPS}} & \multicolumn{1}{l}{biased} & $7.9 \pm 3.3$ & $12.4 \pm 6.0$ & $6.2 \pm 2.6$ & $19.8 \pm 6.6$ \\
                  & \multicolumn{1}{l}{mean} & $0.1 \pm 6.1$ & $-10.9 \pm 7.4$ & $0.2 \pm 2.5$ & $8.8 \pm 8.8$ \\
                  & \multicolumn{1}{l}{no bias} & $5.0 \pm 1.8$ & $3.6 \pm 1.7$ & $4.1 \pm 2.2$ & $11.9 \pm 4.3$ \\ \hdashline
                  
            \multicolumn{1}{p{3.5em}}{\multirow{2}[0]{*}{\parbox{4em}{Cloete \newline{}\etal \cite{Cloete2008}}}} & \multicolumn{1}{l}{biased} & $11.5 \pm 6.4$ & $16.9 \pm 3.6$ & $9.6 \pm 5.1$ & $16.0 \pm 8.8$ \\
                  & \multicolumn{1}{l}{no bias} & $8.5 \pm 5.0$ & $5.8 \pm 3.8$ & $7.3 \pm 5.2$ & $7.9 \pm 4.9$ \\ \hdashline
            \multicolumn{2}{l}{Hu \etal \cite{Hu2015}} & $4.9 \pm 3.5$ & $6.8 \pm 3.0$ & \multicolumn{1}{c}{-} & \multicolumn{1}{c}{-}  \\
            \multicolumn{2}{l}{Tadano \etal \cite{Tadano2013}} & $10.1 \pm 1.0$ & $7.9 \pm 1.0$ & \multicolumn{1}{c}{-}  & \multicolumn{1}{c}{-} \bigstrut[b]\\
            \hline
            \hline
            \multicolumn{2}{l}{Joint Angle CC} & \multicolumn{1}{c}{knee sagittal} & \multicolumn{1}{c}{hip sagittal} & \multicolumn{1}{c}{hip frontal} & \multicolumn{1}{c}{hip transverse} \bigstrut\\
            \hline
            \multicolumn{2}{l}{\emph{CKF-3IMU}} & $0.87 \pm 0.08$ & $0.74 \pm 0.12$ & $0.62 \pm 0.12$ & $0.33 \pm 0.12$ \bigstrut[t]\\
            \multicolumn{2}{l}{\emph{OSPS}} & $0.97 \pm 0.04$ & $0.95 \pm 0.06$ & $0.72 \pm 0.19$ & $0.26 \pm 0.20$ \\
            \multicolumn{2}{l}{Cloete \etal \cite{Cloete2008}} & $0.89 \pm 0.15$ & $0.94 \pm 0.08$ & $0.55 \pm 0.40$ & $0.54 \pm 0.20$ \\ 
            \multicolumn{2}{l}{Hu \etal \cite{Hu2015}} & $0.95 \pm 0.04$ & $0.97 \pm 0.04$ & \multicolumn{1}{c}{-} & \multicolumn{1}{c}{-} \\
            \multicolumn{2}{l}{Tadano \etal \cite{Tadano2013}} & $0.97 \pm 0.02$ & $0.98 \pm 0.01$ & \multicolumn{1}{c}{-} & \multicolumn{1}{c}{-} \bigstrut[b]\\
            \hline
            \hline
            \end{tabular}%
          \label{tab:knee-hip-angle-rmse-cc}%
          % \end{mdframed}
        \end{table}

    \subsection{Total travelled distance (TTD) deviation}
    \changed{
        \emph{CKF-3IMU} was able to estimate global position at $3.6$ - $4.9$\% TTD deviation, which is not as good as state-of-the-art dead reckoning applications \cite{Jimenez2010a, Zhang2017a};
            some of the reason for this also lies in the assumption that the velocity of the shank IMU is zero when the associated foot touches the floor, but of course this IMU continues to move with some small velocity during the stance phase.
        However, pelvis drift has been reduced substantially compared to \cite{Marcard2017}. 
        Note that the experiment was performed in a small room ($4 \times 4$ m$^2$) to capture lower body pose (the main parameter the algorithm aims to solve). 
            Note, dead reckoning algorithms are typically tested on longer courses (\eg{} \texttildelow$125 - 300$ m long, sometimes with different floor levels), which must be consider when interpreting the global positioning results presented here.
    }
    
    \subsection{Pelvis drift and numerical issues}
\begin{figure}[htbp]
            \centering
            \subfloat[Front]{
                \includegraphics[height=4cm]{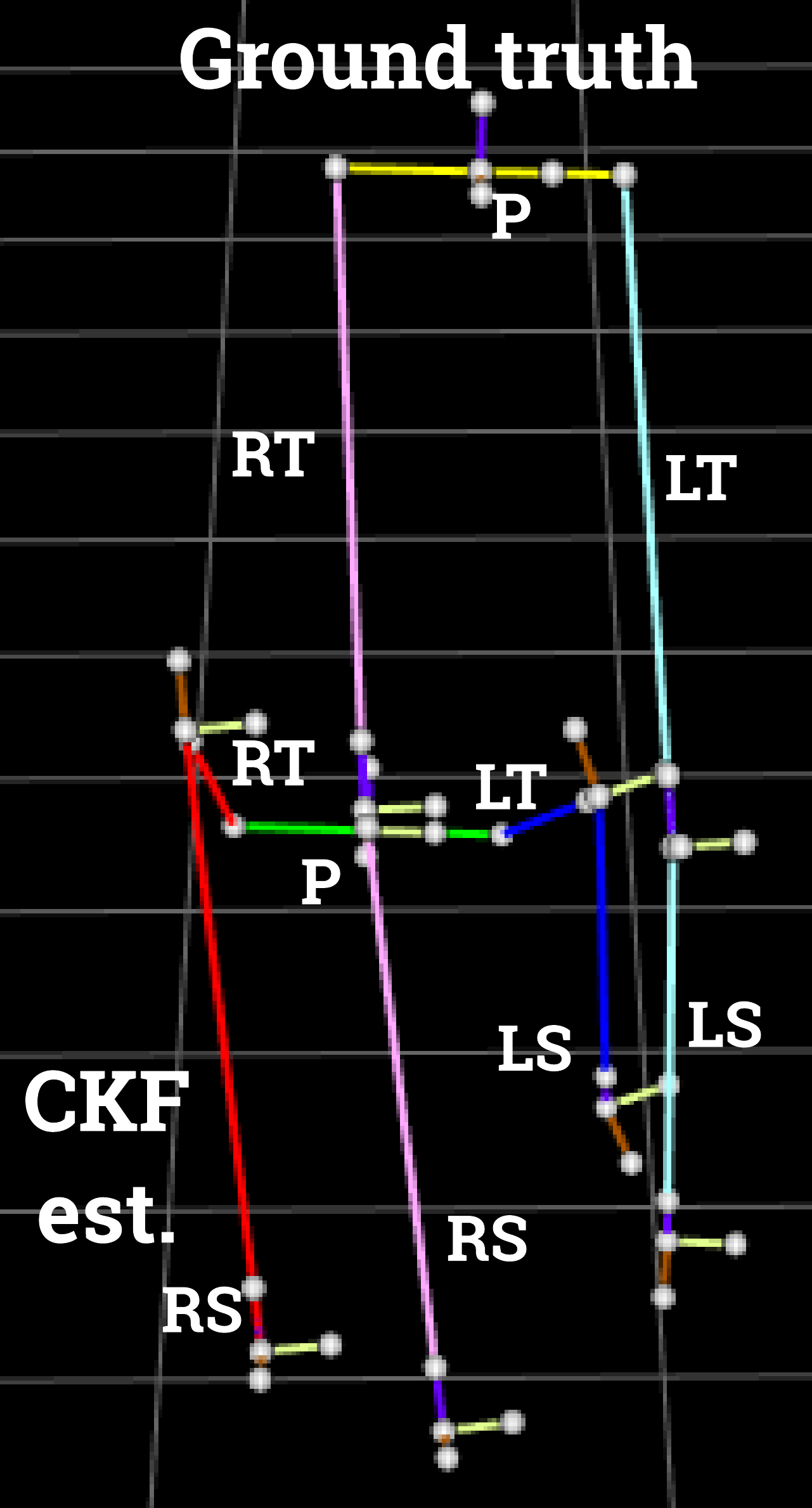}
            }
            \subfloat[Side]{
                \includegraphics[height=4cm]{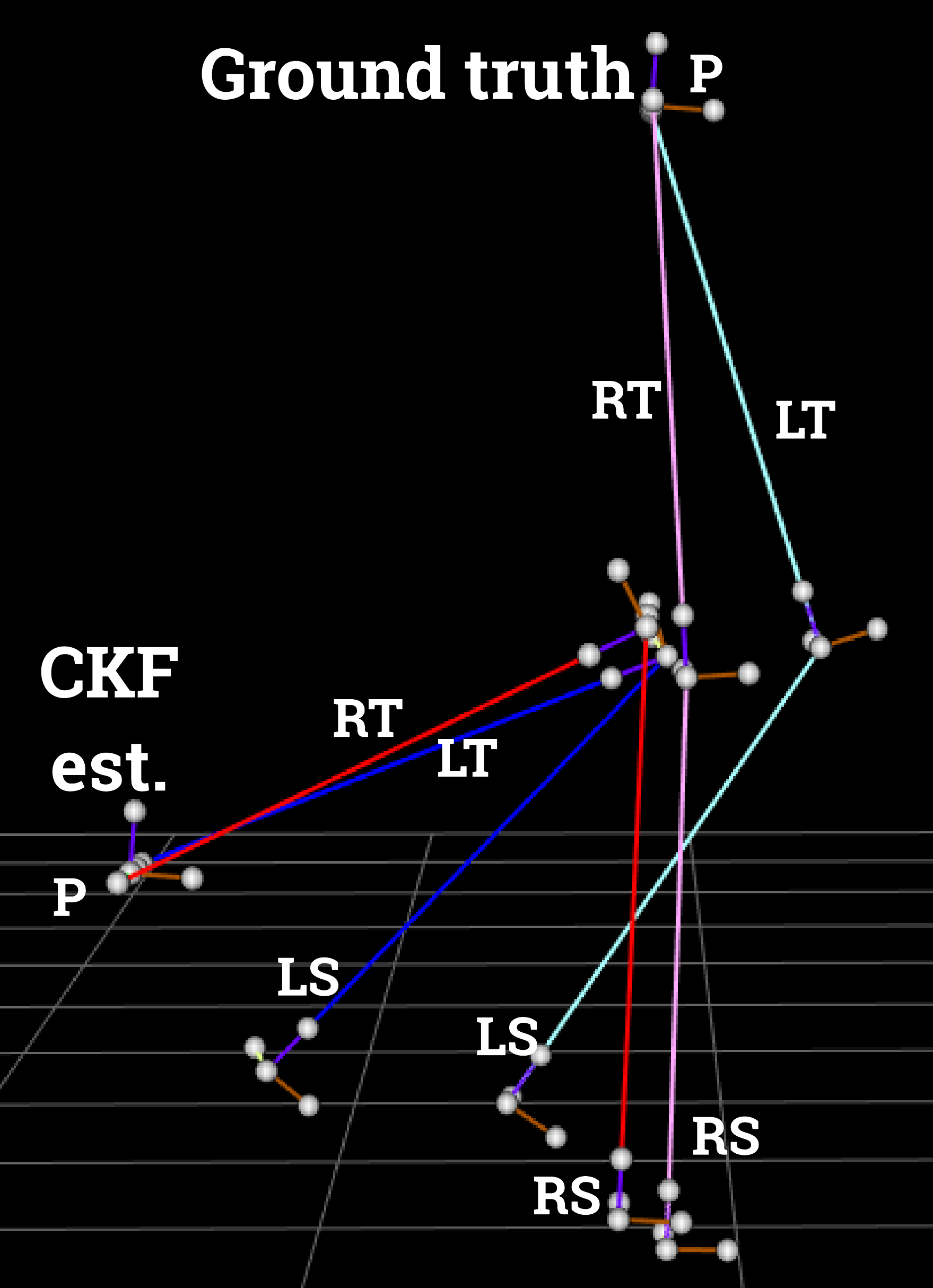}
            }
            \caption{Snapshot of crouching pose estimate (pelvis drift) in approximate front and side perspective views. The red-green-blue and pink-yellow-cyan lines refer to \emph{CKF-3IMU} estimated and ground truth poses, respectively.}
            \label{fig:results-crouchdrift-example}
        \end{figure}

        Without the pelvis position pseudo-measurements during the measurement update of the KF (Section \ref{sec:meas-update}) and without the non-decreasing knee angle during the constraint update (Section \ref{sec:const-nonlin-update}), the resulting pose estimate may drift into a crouching pose as shown in Fig. \ref{fig:results-crouchdrift-example}.
	        % This issue may be similar to the failure that Marcard \etal{} encountered when they tried SIP over a single frame instead of multiple frames \cite{Marcard2017}.
    	    % It is worth mentioning that the flat floor assumption also helped in preventing the crouching pose (divergence) for the walking motions by ensuring that the ankles are not drifting upwards.
	        % Unfortunately, CKF had no sensor measurement describing the knee angle which can help prevent the crouching pose (divergence) and provide better pose estimate.
        
        Without the covariance limiter during the measurement update (Section \ref{sec:meas-update}), the covariance of the CKF algorithm becomes badly conditioned for the motion \emph{5-minute walk} as the position uncertainty grows at a faster rate for the pelvis, as the ankle velocities are zeroed at the time of each footstep, which subsequently limits the rate at which the positional uncertainly of the ankles grows.
            Expectantly, this was not an issue for the shorter walking motions.
            % Alternatively, the issue may be resolved by resetting the covariance based upon certain assumptions every fixed interval of time or when certain conditions are met.
        
        As part of future work, methods to update the state error covariance matrix using information obtained during the constraint update will be explored. 
            In the current framework, the update of the \textit{a posteriori} state error covariance matrix, $\kfcm{P}{k}$, during the constraint step expectedly resulted in rank-deficiency (\ie{} singular) due to the eigenvectors of this covariance matrix being projected onto the constraint surface, which spans a subspace of the state space.
            This rank-deficiency ultimately resulted in pose estimates that are physically impossible or numerically undefined.
        Furthermore, methods which can better track states and covariances of non-linear systems (\eg{} body pose) will also be explored.
            Indeed, this weakness is a well-known limitation of the KF when applied to non-linear system models, but can be better addressed using an unscented KF or a particle filter \cite{Julier2000, simon2006optimal}.
        %Another approach attempted was to avoid singularity is by executing the constraint update only when the covariance matrix is \lq\lq noisy\rq\rq enough (\ie{} nonsingular). 
        %    Although the resulting system was stable, the pose estimate had very high errors as the estimated covariance matrix did not accurately represent the non-linear system (\ie{} body pose).
        Finally, in addition to position and velocity, orientation of the body segments will also be included as state variables to allow the KF to make corrections to segment orientations, compensating for errors in the IMU orientation measurements.
        
    \subsection{CKF limitations: towards long-term tracking of ADL}
        The assumptions in the measurement and constraint update of the CKF algorithm impose limitations to its general usability.
            The pelvis position pseudo-measurements used during the measurement update helps prevent the pelvis drifting away from the feet. 
                However, it will also prevent accurate pose estimation of motions such as sitting and lying down, where the pose is maintained for a duration much longer than that of a typical gait cycle.
                As part of future work, the constraint may be replaced by balance-based constraints that take account of pelvis contact with an external object (\eg{} detect if pelvis has contact with a chair or bed).
            The flat-floor assumption during the measurement update limits the system to environments were the altitude is not changing and is unable to track situations such as walking up ramps or climbing stairs.
            % The hinge knee joint assumption during constraint update also prevents the system from tracking people with varus or valgus deformity, or those capable of hyperextending the knee.
            Lastly, the sensors are assumed rigidly attached to corresponding body segments, whereas in reality sensors are attached using tight velcro straps which may move during dynamic movements like jumping or running.
        
        The CKF algorithm is expected to work on pathological gait where the biomechanical assumptions of the constraint model are not violated. 
            For example, subjects with knee osteoarthritis have demonstrated reduced knee excursion (peak knee flexion angle minus angle at initial contact in degrees) in the sagittal plane \cite{Childs2004}. 
            The CKF should be able to capture such motion given the subject has no varus or valgus deformity (hinge knee joint assumption), even if it includes an atypical spatiotemporal gait pattern; for example, intermittent stopping during the 5-minute walking trial. 
            However, this paper provides an initial proof-of-concept evaluation of the proposed algorithm using an RSC configuration on participants with normal gait and lower body biomechanics. 
            Future work validating the accuracy of this system in a larger and more diverse cohort would help quantify its ultimate clinical utility.
       		
        The CKF algorithm was also highly dependent on the step detection algorithm and the initial calibration step which defines the relative orientation between each IMU and its body segment.
            A step detection error can cause a large velocity and displacement correction on the same ankle at the next detected step which adds to the global position drift.
            
        Of course, in everyday life, the \changed{OMC} reference orientation and acceleration will not be available for the \changed{sensor-to-body} orientation alignment and yaw offset alignment.
            A practical initial or even online calibration procedure that takes into account magnetic interference at the ankle sensors close to the floor will therefore be needed.
            There are different ways of sensor-to-segment calibration in existing literature ranging from static to online/adaptive calibration.
                For example, the subject may be asked to do a predefined posture at the start of pose estimation for orientation alignment \cite{Meng2013, Roetenberg2009}.
                In addition, the subject may be asked to walk in a straight line and then back to the starting point for yaw offset alignment.
                The global yaw correction resulting in the highest $\cs{y}$ axis (sagittal plane) range of motion may be assumed to be aligned with the pelvis anteroposterior axis.
            The magnetic field is known to be inhomogeneous in indoor environments, typically with stronger disturbances closer to the floor \cite{DeVries2009}. 
                Our algorithm would benefit from orientation estimation algorithms which are robust to magnetic disturbances, such as those presented by Del Rosario et al. \cite{DelRosario2018}. 
                Unfortunately, recent advances in automatic sensor-to-segment calibration and magnetometer correction that utilizes two IMUs and a hinge joint may not be applicable to RSC configurations (\eg{} our algorithm only has one IMU on one side of the hinge joint) \cite{Laidig2017a}.
	        % Maybe added
	        % Note that due to having less sensors (\ie{} less measurements), an accurate calibration is even more important as that is the only source of information and there is no other measurement to correct it.
	\section{Conclusion}
	    This paper presented a CKF-based algorithm to estimate lower limb kinematics using a reduced sensor count configuration, and without using any motion database.
	    The mean position and orientation (no bias) RMSEs (relative to the mid-pelvis origin) were $5.21 \pm 1.3$ cm and $16.1 \pm 3.2^\circ$, respectively. 
        The sagittal knee and hip angle RMSEs (no bias) were $10.0 \pm 2.9^\circ$ and $9.9 \pm 3.2^\circ$, respectively, while the CCs were $0.87 \pm 0.08$ and $0.74 \pm 0.12$.
        The performance is acceptable for animation applications. However, accuracy needs to improve for clinical applications.
        To improve performance, additional information is needed to prevent pelvis drift (\eg{} utilize sensors that give pelvis distance, position, or angle of arrival relative to the ankle). 
        % This paper calls the sensor community to develop wearable sensors that can give pelvis information relative to the ankle (\eg{} distance).
        Lastly, the paper contributes a new \changed{OMC (\ie{} Vicon)} and IMU dataset which contains longer duration walking and will be made available at \github{}.

	\begin{comment}
	\section{Authors' Contribution}
	    \begin{itemize}
	        \item Conception of the work/system design`: HK, LK, NL, SJR
	        \item Algorithm design: LS, MR, MDR, SJR
            \item Experiment design: LS, MR, MDR
            \item Data collection: LS
            \item Data analysis and interpretation: LS, MR, MDR, SJR
            \item Drafting/critical revision of the article: LS, MR, MDR, NL, SJR
	    \end{itemize}
    \end{comment}
    
	\section{Acknowledgement}
	    The authors would like to thank Neuroscience Research Australia (NeuRA) for the technical support and access to their Gait laboratory; Lucy Armitage for the crash course on anatomical landmarks.
	    Michael Raitor was supported through a Fulbright scholarship.
	    This research was supported by an Australian Government Research Training Program (RTP) Scholarship.
        The source code for the CKF algorithm and a sample video will be made available at \github{}.
    
    \section*{References}
    \printbibliography[heading=none]
    
\end{document}